\pdfoutput=1
\PassOptionsToPackage{backref=page}{hyperref}
\documentclass[11pt]{article}

\newif\ifarxiv
\arxivtrue %
\newif\ifperfect
\perfecttrue %

\newif\ifshowfig
\showfigtrue %

\newif\ifthesis
\thesisfalse

\ifarxiv
    \usepackage{EMNLP2023}
\else 
    \usepackage[review]{EMNLP2023}
\fi

\usepackage{times}
\usepackage{latexsym}
\usepackage[T1]{fontenc}

\usepackage[utf8]{inputenc}

\usepackage{url}

\makeatletter
\newcommand{\myhref}[2]{\hyper@linkurl{\urlstyle{same}\href{#1}{#2}}{#1}}
\makeatother

\usepackage{sec/package}

\newcommand{\ourdata}{\textsc{Nlp4sgPapers}\xspace}
\newcommand{\ourmodel}{{{\textit{PaperAnalyzer}}}\xspace}

\title{

Beyond Good Intentions: 
\\
Reporting the Research Landscape of 
NLP for Social Good

}

\author{
Fernando Gonzalez\textsuperscript{\rm 1,}\thanks{\hspace{0.1cm} Equal contribution.} { } { }
Zhijing Jin\textsuperscript{\rm 1,2,}\samethanks { } { }
Bernhard Schölkopf\textsuperscript{\rm 1,2}
\\  %
{\bf 
Tom Hope\textsuperscript{\rm 4, 5}
{ } 
Mrinmaya Sachan\textsuperscript{\rm 1,}\thanks{\hspace{0.1cm} Equal supervision.}
{ } \and 
Rada Mihalcea\textsuperscript{\rm 3,}\samethanks
}
\\
\textsuperscript{\rm 1}ETH Zürich
{ }
\textsuperscript{\rm 2}MPI for Intelligent Systems
{ } \textsuperscript{\rm 3}University of Michigan
\\
\textsuperscript{\rm 4}The Hebrew University of Jerusalem
{ }
\textsuperscript{\rm 5}Allen Institute for AI (AI2)\\
\texttt{\{fgonzalez,jinzhi,msachan\}@ethz.ch}
\quad
\texttt{bs@tue.mpg.de}
\\
\texttt{tomh@allenai.org}
\quad
\texttt{mihalcea@umich.edu}
}

\begin{document}
\maketitle

\begin{abstract}

With the recent advances in natural language processing (NLP),
a vast number of applications have emerged
across various use cases. Among the plethora of NLP applications, many academic researchers are motivated to do work that has a positive social impact, in line with the recent initiatives of NLP for Social Good (NLP4SG). However, it is not always obvious to researchers how their research efforts tackle today's big social problems. Thus, in this paper, we introduce 
\ourdata, a scientific dataset with three associated 
tasks that can help identify NLP4SG papers and characterize the NLP4SG landscape by: (1) identifying the papers that address a social problem, (2) mapping them to the corresponding UN Sustainable Development Goals (SDGs), and (3) identifying the task they solve and the methods they use.  
Using state-of-the-art NLP models, we address each of these tasks and 
use them on the entire ACL Anthology, resulting in a visualization workspace that gives researchers a comprehensive overview of the field of NLP4SG.\footnote{ 
\ifarxiv
    We open-source our entire project:
    \begin{itemize}
        \item Website:  \myhref{https://nlp4sg.vercel.app}{nlp4sg.vercel.app}
        \ifperfect
        \item Data: \myhref{https://huggingface.co/datasets/feradauto/NLP4SGPapers}{huggingface/feradauto/NLP4SGPapers}
        \fi
        \item Code:  \myhref{https://github.com/feradauto/nlp4sg}{github.com/feradauto/nlp4sg}
    \end{itemize}
    
\else
    Our code and data have been uploaded to the submission system, and will be open-sourced after acceptance.
\fi
}

\end{abstract}

\section{Introduction}
\begin{figure}[t]
    \centering
    \includegraphics[width=\columnwidth]{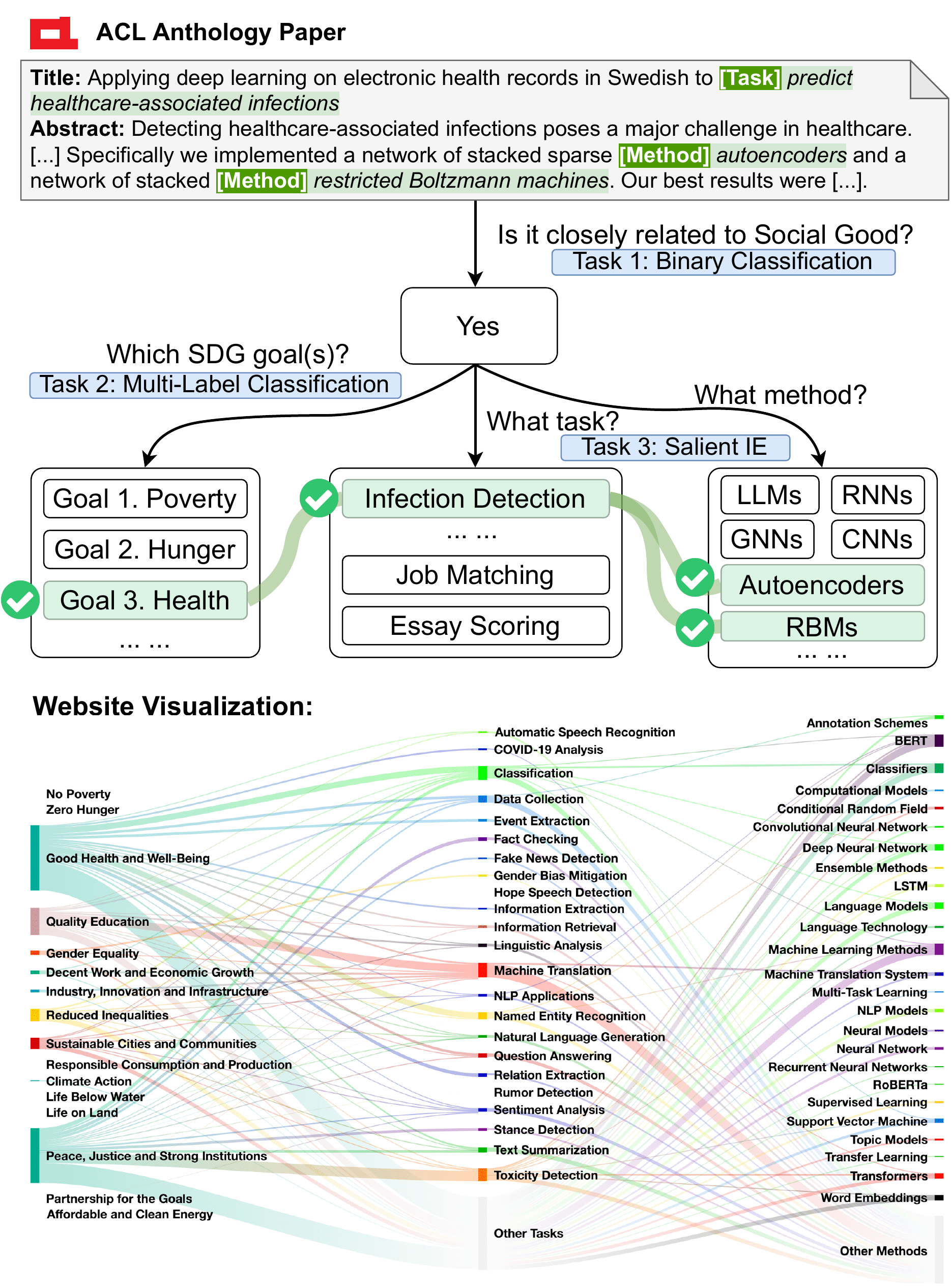}
    \caption{To generate the NLP4SG progress report, we provide a \ourmodel, which identifies social good papers (Task 1), classifies the relevant UN SDGs (Task 2), and analyzes salient scientific terms (Task 3). We process all ACL Anthology papers to create a Sankey diagram (bottom) on our website.
    }
    \label{fig:overview}
    \ifarxiv \else \vspace{-8mm} \fi
\end{figure}

With the rapid advancement of natural language processing (NLP) technology \cite{radford2018improving,devlin-etal-2019-bert,gpt3,ouyang2022instructGPT},  there has been a growing interest in using NLP for applications with social impact in recent years \cite{wu2020deep,mello2019text,jin2023natural}. This effort has been condensed into the initiative of ``NLP for Social Good'' (NLP4SG) \cite{jin-etal-2021-good}, which 
began with an overall theoretical framework \cite{jin-etal-2021-good} in 2021, and is reaching broader community impact through 
various workshops at NLP conferences \cite{nlp4posimpact-2021-nlp,nlp4posimpact-2022-nlp}. 

However, most of the efforts so far have consisted of disparate community discussions, opinion pieces, and  projects, yet without a unifying community-based insight. 
This is partly due to a lack of understanding of how NLP research efforts align with  social good implications.
We argue that it is critical to have  community reports based on \textit{solid empirical evidence}, similar to the United Nations' (UN) annual compilation of the report on Sustainable Development Goals (SDGs) 
\cite[e.g.,][]{un2021sustainable,un2022sustainable}. Such progress reports for NLP4SG can help calibrate {intuitions} with {facts} and  (1) inspire the NLP community to understand how well NLP research address social good; (2) raise awareness of under- or un-addressed areas; (3) serve as a reference for researchers who start in the field when deciding on what topics to take on; and (4) foster better connections between the research community and social science experts, allowing for more impactful research and feedback to improve NLP4SG.

To this end, we design a new suite of tasks for analysing NLP4SG papers, spanning from identification of NLP4SG papers, classifying them into SDGs, to identifying the task and methods used in them. We create \ourdata, a scientific dataset of 5,000 papers annotated with the above three tasks, and use the dataset to build an ``NLP for NLP'' \ourmodel system  (\cref{fig:overview})
that applies the state-of-the-art NLP models to the NLP papers.

Using our \ourmodel system, we parse the entire database of 76K ACL Anthology papers, and find the following trends: (1) 13.07\% of the papers are related to social good; (2) healthcare and education are popular goals among NLP4SG papers, but goals such as poverty and hunger are largely unaddressed; (3) most common tasks addressed by the community in NLP4SG are machine translation, text classification, and {toxicity detection}, and although the most common models are large language models (LLMs) \cite[e.g.,][]{radford2018improving,devlin-etal-2019-bert,liu2019roberta}, these innovations have not been deployed in all tasks.

In summary, our contributions are as follows:
\begin{enumerate}
    \item We introduce a suite of novel tasks to analyze NLP4SG papers, and support it with \ourdata{}, a scientific dataset annotated with 5,000 NLP papers.
    \item We provide our \ourmodel system, which consists of LLM-based models on the three tasks.
    \item We analyze the entire ACL Anthology to understand key trends in NLP4SG from {1980}.
    \item We open-source our NLP tools, and build a publicly accessible website to visualize the analysis results, making the progress tracking accessible for anyone interested in NLP4SG.
\end{enumerate}

\section{Goals of the NLP4SG Report}

We begin by listing the key questions that drive our NLP4SG progress report and the desired properties of a good NLP4SG report, and highlight what purpose the report will serve. Finally, we introduce our proposed NLP-for-NLP pipeline.

\myparagraph{Key Questions.}
When envisioning the NLP4SG progress report, we find the following questions crucial to address:
\begin{enumerate}[label=Q\arabic*]
    \item {\bf Overall estimate.} How large is the proportion of NLP4SG papers among all NLP papers? How does this proportion change over the years? \label{item:q1}
    \item {\bf Social good aspects.} Among NLP4SG papers, what specific social good aspects do they tend to address?\label{item:q2}
    \item {\bf Technical solutions.} For each social good goal, how are the papers addressing them? For example, what tasks do people formulate, and what models do people use? Is there an overlooked space?\label{item:q3}
\end{enumerate}

\myparagraph{Properties of a Good Progress Report.}
After settling in the key questions that the report should aim to answer, we propose several criteria for a good automation of the NLP4SG progress report. A good report should ({\bf P1}) comprehensively address the questions; ({\bf P2}) base its answers on representative, large data; ({\bf P3}) require little repetitive manual effort, ideally largely automated; ({\bf P4}) be transparent and open-sourced so that it is trustworthy; ({\bf P5}) be reader-friendly for both technical and non-technical audiences; and ({\bf P6}) be easy to access for all parties, including but not limited to researchers, social good organizations such as NGOs, funding agencies, and the general public.

\myparagraph{Missions.}
This report aims to serve several purposes. First, we want to convey an overall positive and encouraging message that there is a growing and diverse community of researchers who are interested in NLP4SG. 
Second, we want to improve the cohesion of this community, by making the information on the existing research landscape easier to access, so that people can build on each other's research.
Third,  we aim to inspire more people to develop novel tasks and use cases to fill in the under-addressed social impact areas.
Finally, we want to make the current progress transparent and public to all interested parties, especially non-research parties, including organizations such as NGOs working on NLP4SG, decision-makers, and the general public. By doing this, we can better connect the research community with all other related communities, not only to broadcast the impact of NLP4SG to the broader community, but also to help improve NLP research with external feedback, and formulate research questions more aligned with real-world problems.

\begin{figure}[t]
    \centering
    \includegraphics[width=\columnwidth]{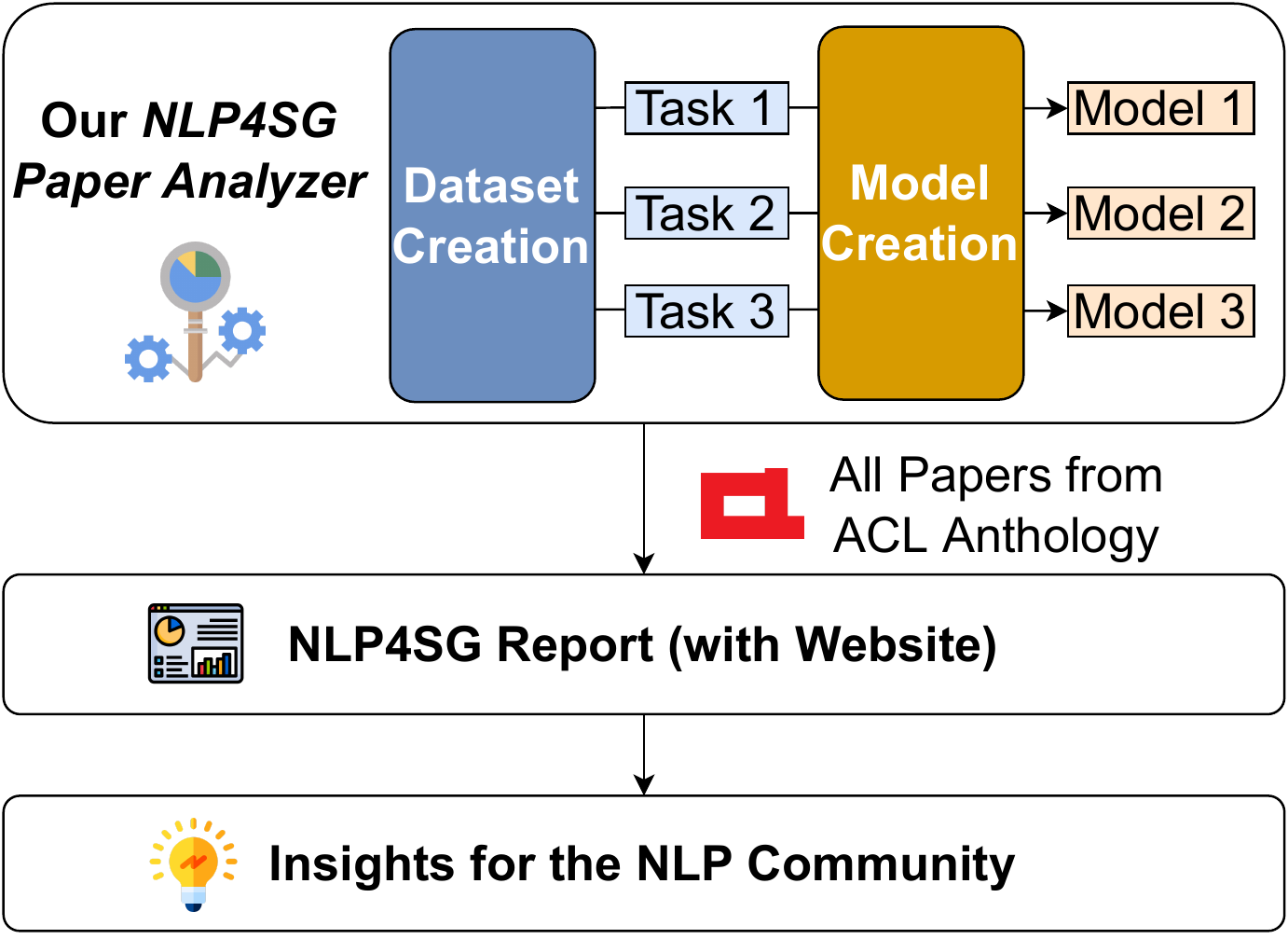}
    \caption{Our pipeline to automate the NLP4SG report.}
    \label{fig:pipeline}
\end{figure}

\myparagraph{Our Proposed NLP-for-NLP pipeline.}
We develop an NLP-for-NLP pipeline to automatically analyze papers for the NLP4SG report, as shown in \cref{fig:pipeline}. As part of this pipeline, we develop a \ourmodel\ system consisting of various models trained to answer the three questions (Q1-Q3).%

The pipeline consists of the following steps: %
(1) We annotate various datasets that can help us answer the three questions (meeting P1 and P2). (2) We use this data to train and evaluated our \ourmodel models.
(3) Next, we run our NLP4SG \ourmodel (which meets P3) on a large repository of NLP papers, and compose a comprehensive NLP4SG progress report, together with an open-source GitHub release of our code (which meets P4) and an interactive interface whose screenshots are in \cref{appd:screenshots} (which meets P5 and P6).
(4) Finally, we distill some key findings for the NLP community, to pave the way for more cooperative, evidence-guided community efforts to fuel NLP4SG.

\section{\ourdata{} Dataset}

We build a \ourdata{} dataset to address the following three tasks as illustrated in \cref{fig:overview}:
\begin{itemize}[topsep=1ex,itemsep=-0.5ex]
    \item \textbf{Task 1:} Binary classification of NLP4SG papers (for \ref{item:q1}).
    \item \textbf{Task 2:} Multi-label multi-class categorization of NLP4SG papers into the 17 UN SDGs (for \ref{item:q2}).
    \item \textbf{Task 3:} Identification of salient scientific terms (tasks and methods)
    in NLP4SG papers (for \ref{item:q3}).
\end{itemize}

Next, we introduce the data collection process, and then analyze the data statistics for each task.

\subsection{Dataset Collection}
\paragraph{Data Source and Preprocessing.}

We identify the ACL Anthology
as an ideal data source. 
We obtain 76,229 papers from ACL Anthology that were published by May {27}, 2022.\footnote{\url{https://aclanthology.org/anthology+abstracts.bib.gz}} We randomly sample 5,000 papers uniformly over the entire ACL Anthology and parse their titles and abstracts. See \cref{appd:data_source,appd:data_preprocessing} for discussions about the data sources and our data preprocessing details.
We split the dataset into 2,500 train samples, 500 development samples, and 2,000 test samples. Statistics of the data are shown in \cref{tab:sg_binary_example}.

\paragraph{Task Formulation.}
Let us denote our dataset as $\bm{D}:=\{\bm{x}_i\}_{i=1}^N$ consisting of $N$ papers, each paper $\bm{x}:=(\bm{t}, \bm{a})$ contains a title $\bm{t}$ and abstract $\bm{a}$. 

Task 1 is a binary classification task $f_1: \mathcal{X} \rightarrow \mathcal{Y}$, which maps each paper to a binary label space $\mathcal{Y} = \{0,1\}$ representing whether a paper address a social good task ($y=1$) or not ($y=0$).

Task 2 is a multi-label multi-class classification task $f_2: \mathcal{X}' \rightarrow \mathcal{Z}$, which maps NLP4SG papers $\mathcal{X}' = \{\bm{x} | f_1(\bm{x})=1, \bm{x} \in \bm{D}\}$ to a space $\mathcal{Z}$, which is the power set of all 17 UN SDGs. 

In Task 3, we extract the main NLP tasks addressed and methods used in the NLP4SG papers, $f_3: \mathcal{X'} \rightarrow \mathcal{W}_{\mathrm{t}} \times \mathcal{W}_{\mathrm{m}}$, where $\mathcal{W}_{\mathrm{t}}$ and $\mathcal{W}_{\mathrm{m}}$ represent the space of the power sets of the main tasks and methods in NLP4SG papers, respectively. Given %
our desiderata, it is better to make the extracted terms easy to use for summarizing trends in our Sankey diagram (\cref{fig:overview}). Inspired by the annotation scheme of Papers with Code,\footnote{\url{https://paperswithcode.com/}} we use task names such as ``machine translation'' and ``toxicity detection''; and we use method names such as ``BERT-based models,'' ``long short-term memory networks (LSTMs),'' and ``support vector machines (SVMs).'' We provide the detailed list in \cref{appd:annot_3}. This allows us to use Task 3 to provide evidence that can help us answer high-level questions such as what NLP4SG work uses  what type of technology, as shown in \cref{fig:overview}.

\paragraph{Data Annotation.}
The data associated with each task was annotated by two NLP researchers who are proficient in English. We provide detailed annotation guidelines for each task in \cref{appd:annotation_guide}. 
Briefly, the inter-annotator agreements are 92.93\% Cohen's kappa for Task 1, 88.67\% Cohen's kappa for Task 2, and achieve a high similarity score of 84.63\% BERTScore \cite{zhang2020bertscore} between the annotator's responses for Task 3. 
We resolved cases of disagreement by first asking them to review potential oversights, and then clarifying the annotation principles for samples that are ambiguous. With these clarifications, the annotators converged to an agreement through re-iteration and discussions.

\subsection{Dataset Characteristics}

\paragraph{Data Overview for Task 1.} We show the overall statistics in \cref{tab:sg_binary_example}. We can observe that 11.84\% of the papers are related to social good (592 papers of our sample of 5000). %

\begin{table}[h]
    \centering
      \resizebox{\columnwidth}{!}{%
    \setlength\tabcolsep{3pt}
    \begin{tabular}{lccccc}
    \toprule
     & \textbf{\# All Papers} & \textbf{\# SG Papers} & \textbf{\# Sents} & \# \textbf{Tokens} & \textbf{Vocab}  \\ \bottomrule
\textbf{Total} & 5,000 & 592
(11.84\%) 
& 40,533 & 656,939 & 45,088 \\ \hline
Train & 2,500 & 284 
& 20,305 & 330,618 & 29,274 \\
Dev & 500 & 57 
& 4,112 & 65,705 & 10,511 \\
Test & 2,000 & 251
& 16,116 & 260,616 & 24,913 \\
    \bottomrule
    \end{tabular}
    }
    \caption{Statistics for Task 1. We show the number of social good-related papers (\# SG Papers), along with the general statistics, including the total number of papers (\# All Papers), sentences (\# Sents), tokens (\# Tokens), and vocabulary size (Vocab).}
    \label{tab:sg_binary_example}
\end{table}

\paragraph{Data Overview for Task 2.}
For Task 2, we first show in \cref{tab:unsdg_distr} the distribution of all 17 UN SDGs among all the 592 social good-related papers. The distribution is strongly skewed, with some goals covering the majority of the data. For example, the top three goals --  G3 Health (34.12\%), G16 Peace (31.93\%), and G4 Education (16.22\%) -- cover a total of 82.27\% of the NLP4SG papers. Many goals rarely occur or are even not present in the dataset, such as G1 Poverty and G2 Hunger.
This distribution suggests either a lack of attention from the NLP4SG community to some of these UN goals or challenges in addressing these goals through NLP. Additional analyses are presented in \cref{sec:insights}.

To enrich the data for low- or zero-occurrence classes, we perform upsampling to increase the number of samples corresponding to all the classes that constitute less than 5\% of our dataset. 
Specifically, we take the unannotated part of the ACL Anthology, run the best model for Task 1 to identify social good papers, and use SDG-specific keywords to obtain highly possible paper candidates for the under-addressed SDGs. Then, we manually check the SDGs that these papers address. By this procedure, we enrich the original data by 167 more papers, which are more distributed over the low- or zero-occurrence classes. Details of the upsampling procedure are in \cref{appd:upsampling}.

\begin{figure}[ht]
    \centering
    \includegraphics[width=\columnwidth]{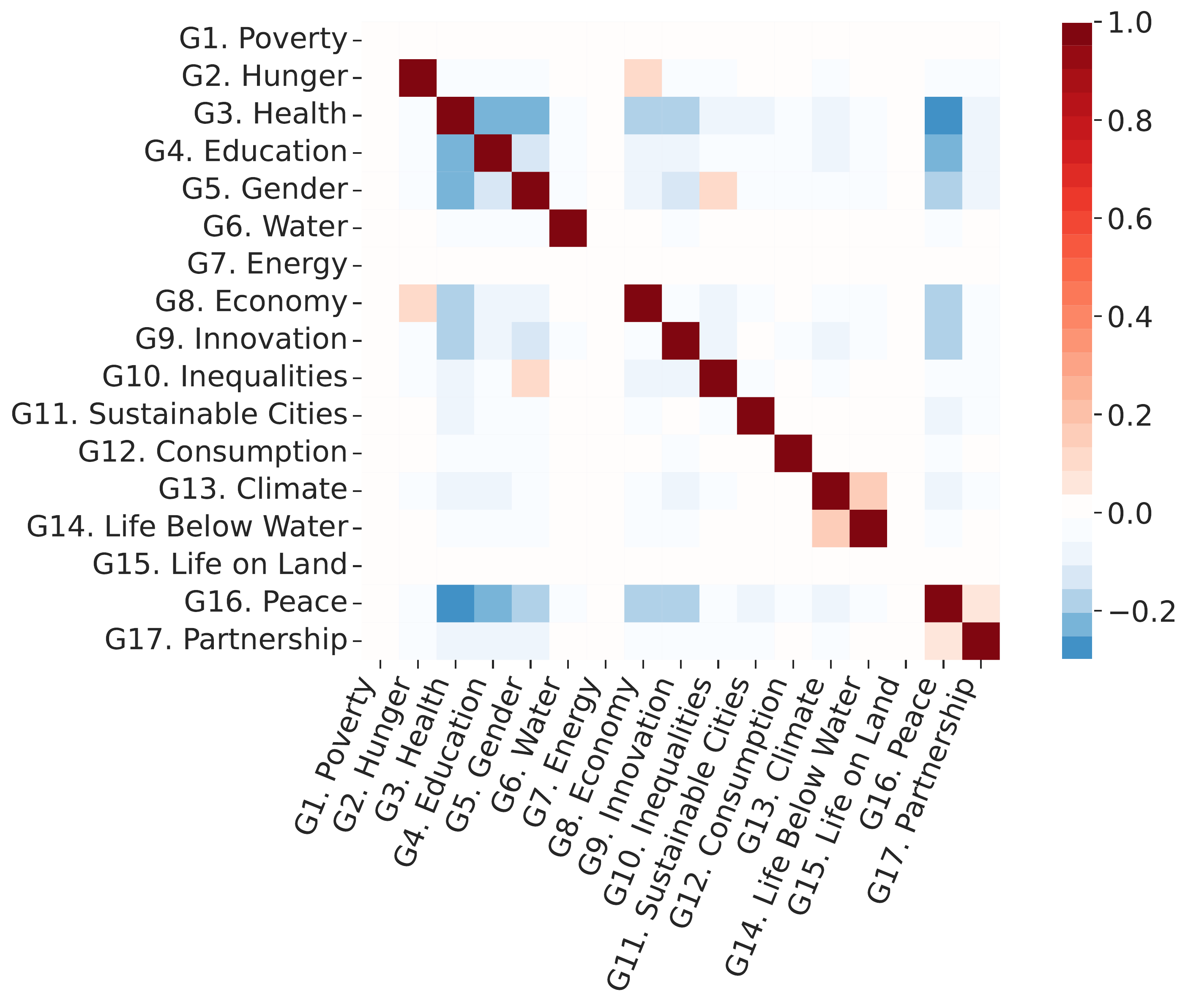}
    \caption{The correlation heatmap of the 17 UN SDGs.}
    \label{fig:unsdg_heatmap}
    \ifarxiv \else
    \vspace{-6mm} \fi
\end{figure}

We report the statistics of the upsampled dataset in \cref{tab:unsdg_distr} and more in \cref{appd:upsampling}.
Since each paper can correspond to several UN SDGs (on average, 1.10 goals per paper), we plot the heatmap of co-occurrences of the SDGs in \cref{fig:unsdg_heatmap}. As we can see, some of the top correlations are seen between G2 Hunger and G8 Economy, as well as between G5 Gender and G10 Inequalities.

\begin{table}[ht]
    \centering\small
     \begin{tabular}{lccccc}
\toprule
\multirow{2}{*}{\textbf{UN SDG}} & \multirow{2}{*}{\textbf{Proportion}} & \textbf{\# Samples} \\
& & {(Before $\rightarrow$ After)} \\ \midrule
G1. Poverty & 0.00\% & 0 $\rightarrow$ 0  \\
G2. Hunger &0.00\% & 0 $\rightarrow$ 4 \\
G3. Health & 34.12\% & 202 $\rightarrow$ 202  \\
G4. Education & 16.22\% & 96 $\rightarrow$ 97  \\
G5. Gender & 3.04\% & 18 $\rightarrow$ 85  \\
G6. Water & 0.00\% & 0 $\rightarrow$ 1  \\
G7. Energy &0.00\% & 0 $\rightarrow$ 0 \\
G8. Economy &2.36\% & 14 $\rightarrow$ 58 \\
G9. Innovation & 13.85\% & 82 $\rightarrow$ 97  \\
G10. Inequalities & 5.24\% & 31 $\rightarrow$ 32  \\
G11. Sustainable Cities & 0.84\% & 5 $\rightarrow$ 16  \\
G12. Consumption & 0.17\% &  1 $\rightarrow$ 1  \\
G13. Climate & 0.17\% & 1 $\rightarrow$ 20  \\
G14. Life Below Water &0.00\% & 0 $\rightarrow$ 2 \\
G15. Life on Land & 0.00\% & 0 $\rightarrow$  0  \\
G16. Peace & 31.93\% & 189 $\rightarrow$ 201  \\
G17. Partnership & 2.70\% & 16 $\rightarrow$ 21  \\
\bottomrule
    \end{tabular}
    \caption{Statistics of the data for Task 2. We first show the natural proportion of UN SDGs in the dataset, which motivates us to do upsampling. And we show the number of samples before and after 
    upsampling. G1, G7, and G15
    remain absent from the dataset even after upsampling.
    }
    \label{tab:unsdg_distr}
\end{table}

\paragraph{Data Overview for Task 3.}

Since there has been significant previous work \citep{jain-etal-2020-scirex,luan-etal-2018-multi} on extracting salient information from scientific literature, we only annotate test data for this task to evaluate existing models in our domain of NLP4SG.
On average, the spans of task terms are {3.6} words long, and the spans of method terms have {2.4} words long.
Also, each paper has an average of {1.06} tasks and {1.5} methods, and the entire set has a total of {263} unique tasks and {335} unique methods.

\section{NLP4SG \ourmodel}

Our \ourmodel system works as shown in \cref{fig:overview}, with three sequential steps: we first train a binary classifier (addressing Task 1), followed by a UN SDG goal classifier (addressing Task 2), and finally processing the paper with a pretrained task and method extractor (addressing Task 3).

\subsection{NLP4SG Binary Classifier}\label{sec:model_task1}
\paragraph{Experimental Setup.}
We adopt several commonly used LLMs on our \ourdata{} dataset, including BERT \cite{devlin-etal-2019-bert}, RoBERTa \cite{liu2019roberta}, SciBERT \cite{beltagy2019scibert}, and InstructGPT \cite{ouyang2022instructGPT} text-davinci-002.

We use distant supervision (DS) for data augmentation when finetuning the models. To obtain distantly supervised data, we use two methods to identify additional NLP4SG papers from the unannotated set of almost 76K papers.
We first apply keyword matching using a list of curated NLP4SG keywords such as ``healthcare'' and ``education'', and then we also include papers with high cosine similarity with the textual descriptions of the UN SDGs. See the full list of keywords and details of how we compile the distantly supervised data in \cref{appd:augment}, and see our model implementation details in \cref{appd:model1_implementation}.

\begin{table}[ht]
    \centering \small
    \setlength\tabcolsep{4pt}
    \begin{tabular}{lcccc}
    \toprule
    & F1 & Acc & P & R \\ \midrule
    Random (Proportional)     & 6.68 & 79.05 & 7.61 & 5.95 \\
    Random (Uniform) & 15.65 & 46.65 & 9.77 & 39.29 \\
    InstructGPT (Zero-Shot) & 28.83 & 80.25  &  27.49 & 30.30 \\
    InstructGPT (Few-Shot) & 40.72 & 86.75 & 49.73 & 34.47 \\
    BERT & 65.87  &  92.90 &  83.03  &  54.58 \\
    BERT + DS & 73.14  &  94.05 &  84.38  &  64.54\\
    RoBERTa & 74.89 & 94.30 & 83.74 & 67.73 \\
    RoBERTa + DS  & 75.16 & \textbf{94.35} & 83.82 & 68.13 \\
    SciBERT & 73.42 & 94.10 & \textbf{84.90} & 64.68 \\
    SciBERT + DS & \textbf{75.98} & 94.15 & 78.39 & \textbf{73.71} \\ 
    \bottomrule
    \textbf{Ablation Study} \\
   SciBERT + DS by Similarity & 62.96 & 89.35 & 55.86 & 72.11 \\
   SciBERT + DS by Keyword  & 74.68  &  94.10 & 80.93  & 69.32\\
    \bottomrule
    \end{tabular}
    \caption{Model performance on Task 1, including the random baselines, whose distribution are proportional to the label distribution or uniform; as well as common LLMs, with and without our proposed DS technique. We report the F1 score, precision (P), and recall (R) of the positive class, as well as accuracy (Acc). 
    For the best-performing model, SciBERT+DS, we also conducted ablation studies removing the different subsets of our DS data.
    }
    \label{tab:res_task1}
\end{table}
\myparagraph{Results.}
We show our model performance for the binary classification task in \cref{tab:res_task1}. The best-performing model is SciBERT+DS, which is enhanced by the addition of our DS technique. DS contributes 2.56 percentage points (pp) improvement over the baseline SciBERT model on fully supervised data only. 
Moreover, with the ablation study, we show that by using only DS by keyword we get an improvement of 1.26 pp on the F1 score. Using only similarity DS does not improve the F1 score, but causes an increase in recall of 7.43 pp. Both DS techniques combined help the model to increase the F1 score to 75.98\%. We choose the model with the best F1 score because we want to get a precise but also a complete report of the NLP4SG landscape.

\myparagraph{Interpretability \& Error Analysis.}
To understand the working of our model, we apply a commonly used interpretability tool, the local interpretable model-agnostic explanations (LIME) \cite{ribeiro-etal-2016-trust} to our best performing model, SciBERT+DS. LIME outputs the words that our model relies on as positive and negative evidence for the classification. Some common positive keywords for our model include ``biomedical'', ``students'', and ``social'', while some negative keywords are ``lexical'', ``parsing'' and ``grammar''. 
The details of our LIME implementation and more visualized examples can be found in \cref{ap_res_t1}.

Our model makes a few common error types, including: (1) For false positives (FP), many papers include the common word ``social'', which appears in 10.10\% of the FP papers. We suspect that this is due to the frequent use of ``social media'', which is not necessarily related to social good, and the fact that ``social'' is a good indicator of an NLP4SG paper since it appears in 15.03\% of the true positive papers.
(2) Among false negative (FN) samples, a large subset is innovation and education-related papers, constituting 30.30\% and 27.27\%  respectively among the FN, larger than their proportion (13.55\% and 23.90\%) in the entire test set. A reason might be that the frequent use of words like ``learning'' and 
scientific vocabulary makes our model struggle identifying a research paper that uses machine learning vs one addressing education topics.

\subsection{UN SDG Goal Classifier}\label{sec:model_task2}

\paragraph{Experimental Setup.} 
For Task 2, we evaluate zero-shot models using all the 759 annotated NLP4SG papers as our test set, as it is challenging to train classifiers for 17-class classification with large imbalance that results in several low-occurrence classes. Specifically, we first adopt the pretrained MNLI classifiers, including BART   \cite{lewis-etal-2020-bart}, DistilBERT \cite{sanh2019distilbert}, and DeBERTa \cite{he2021deberta}. We also include the InstructGPT model \cite{ouyang2022instructGPT}, which we ask to generate a list of relevant SDGs for the paper. More experimental setup details are in \cref{appd:model2_implementation}, including the prompts, implementation details for the models, and the evaluation metrics.

\myparagraph{Results \& Error Analysis.}
We evaluate all models and report their results in \cref{tab:res_3_micro}. Among all the models, InstructGPT has the highest performance, scoring 69.37\% F1. It also has the highest performance on all other measures, such as partial list match with 74.57\% of the samples, and exact list match with 66.66\% of the samples.

\begin{table}[ht]
    \centering \small
    \setlength\tabcolsep{4pt}
    \begin{tabular}{lcccccc}
    \toprule
         Model & F1 & PM & EM & P & R \\ \midrule
         BART & 27.30 &37.55 &10.80 & 44.22 & 35.65 \\
         DistilBERT & 26.09 & 58.23 & 3.95 & 17.78 & 56.66 \\ %
         DeBERTa & 25.50 & 39.13 & 11.20 & 37.82 & 35.89 \\ %
         InstructGPT  & \textbf{69.37} & \textbf{74.57} & \textbf{66.66} & \textbf{77.64} & \textbf{67.95} \\
         \bottomrule
    \end{tabular}
    \caption{Model performance on Task 2. 
    We report the weighted F1, percentage of samples whose ground-truth list of goals exactly matches (EM) with the model output, percentage of samples whose ground-truth list of goals partially matches (PM) with the model output, precision (P), and recall (R). 
    }
    \label{tab:res_3_micro}
\end{table}

\begin{table}[ht]
    \centering\small
    \setlength\tabcolsep{4pt}
    \begin{tabular}{lccc}
\toprule
{UN SDG} 
& F1 & P  &  R   \\ \midrule
G1. Poverty & -- & -- & --  \\
G2. Hunger & 75.00 & 75.00& 75.00 \\
G3. Health & 88.78 & 87.50 & 90.10 \\
G4. Education & 73.45 &   64.34  &  85.57   \\
G5. Gender & 84.56  &  98.44 &  74.12   \\
G6. Water & 100.00 & 100.00 & 100.00    \\
G7. Energy & -- & -- & --  \\
G8. Economy & 54.32 & 95.65  &  37.93  \\
G9. Innovation & 26.45  &  66.67   & 16.49  \\
G10. Inequalities &  27.40 &   24.39 &   31.25   \\
G11. Sustainable Cities &  44.78 &   29.41  &  93.75   \\
G12. Consumption & 50.00 & 33.33 & 100.00 \\
G13. Climate & 85.71 & 100.00 & 75.00  \\
G14. Life Below Water & 80.00 & 66.67 & 100.00  \\
G15. Life on Land & -- & -- & --  \\
G16. Peace & 80.74  & 84.07 &  77.66   \\
G17. Partnership & 0.00& 0.00&0.00  \\ \hline
Weighted Average & 69.37 & 77.64 & 67.95 \\
\bottomrule
    \end{tabular}
    \caption{The class-specific performance of our best-performing InstructGPT method.} 
    \label{tab:res_task3}
    \ifarxiv \else
    \vspace{-4mm} \fi
\end{table}

We also perform a more detailed breakdown of the performance of our best-performing model by different UN SDGs in \cref{tab:res_task3}. We can see that it is easy for the model to distinguish goals such as health, gender equality, and climate. However, other goals such as social inequalities and innovation are ambiguous for the model. One possible reason is that some NLP papers use domain-specific words such as ``{bias},'' which can also mean data bias, or spurious correlations in the data, so the model might confuse it with the general meaning of bias towards or against certain groups of people in society.

\subsection{Task and Method Analyzer}\label{sec:model_task3}
\paragraph{Experimental Setup.}
Although our goal is to extract the main tasks and methods for our Sankey diagram, our setting is novel and distinct from traditional sentence-level named entity recognition (NER), which aims at span-level extraction of all named entities. Instead, our setting is document-level as we just need one mention of the same term in case there are co-references or paraphrases. We also accept generative answers if there are no appropriate existing spans.
Thus, we evaluate Task 3 using common metrics for extractive and generative question answering (QA) \cite{rajpurkar2016squad,chen-etal-2019-evaluating,sai2023survey}, such as exact match, F1, and BERTScore \cite{zhang2020bertscore}.

We first apply commonly used LLMs finetuned on QA datasets, such as BERT \cite{devlin-etal-2019-bert}, RoBERTa, and DeBERTa \cite{he2021deberta}. We also include the scientific NER models, PURE \cite{zhong-chen-2021-frustratingly} and SciREX \citep{jain-etal-2020-scirex}. Finally, we include the general-purpose model, InstructGPT \cite{ouyang2022instructGPT}. Model implementation details are in \cref{appd:model3_implementation}.

\begin{table}[ht]
    \centering\small
    \setlength\tabcolsep{2pt}
     \resizebox{\columnwidth}{!}{%
    \begin{tabular}{l
    ccc
    @{\extracolsep{7pt}}
    ccc
    }
    \toprule

\multirow{2}{*}{Model} & \multicolumn{3}{c}{Task}  & \multicolumn{3}{c}{Method}  \\
\cline{2-4} \cline{5-7}
& BS &  F1 & EM & BS & F1 & EM \\ 
\midrule
BERT  & 82.29 &  35.08 & 15.94 & 75.04 &  23.08 & 4.38\\
RoBERTa  & 81.39 & 33.43 & 14.74 & 76.47 &  27.30 &9.56\\
DeBERTa & 80.36 &  31.28 & 11.16 & 75.62 &  26.05 & 4.78 \\
PURE & 71.30  &    34.67   & 5.58   & 69.65      &  30.43 &5.98\\
SciREX &  84.30 &   \textbf{49.03}   &  14.74  & 72.14     & \textbf{33.70}  &  6.37\\
InstructGPT & \textbf{84.65}  & 43.94 & \textbf{19.12} & \textbf{77.95} &  32.05 & \textbf{11.55} \\

    \bottomrule
    \end{tabular}
     }
    \caption{Model performance on Task 3 by BERTScore F1 (BS), F1, and Exact Match (EM). Due to space limitations, we omit the ``text-davinci'' when describing the InstructGPT.
    }
    \label{tab:ie_tasks}
\end{table}

\myparagraph{Results.} 
We show the model performance for Task 3 in \cref{tab:ie_tasks}. The best model by BERTScore is InstructGPT-002, reaching {84.65}\% for task extraction, and {77.95}\% for method extraction. It also achieves a higher exact match score than other strong models such as SciREX.

\section{Insights}\label{sec:insights}

\begin{figure}[ht]
    \centering
    \ifshowfig
    \includegraphics[width=\columnwidth]{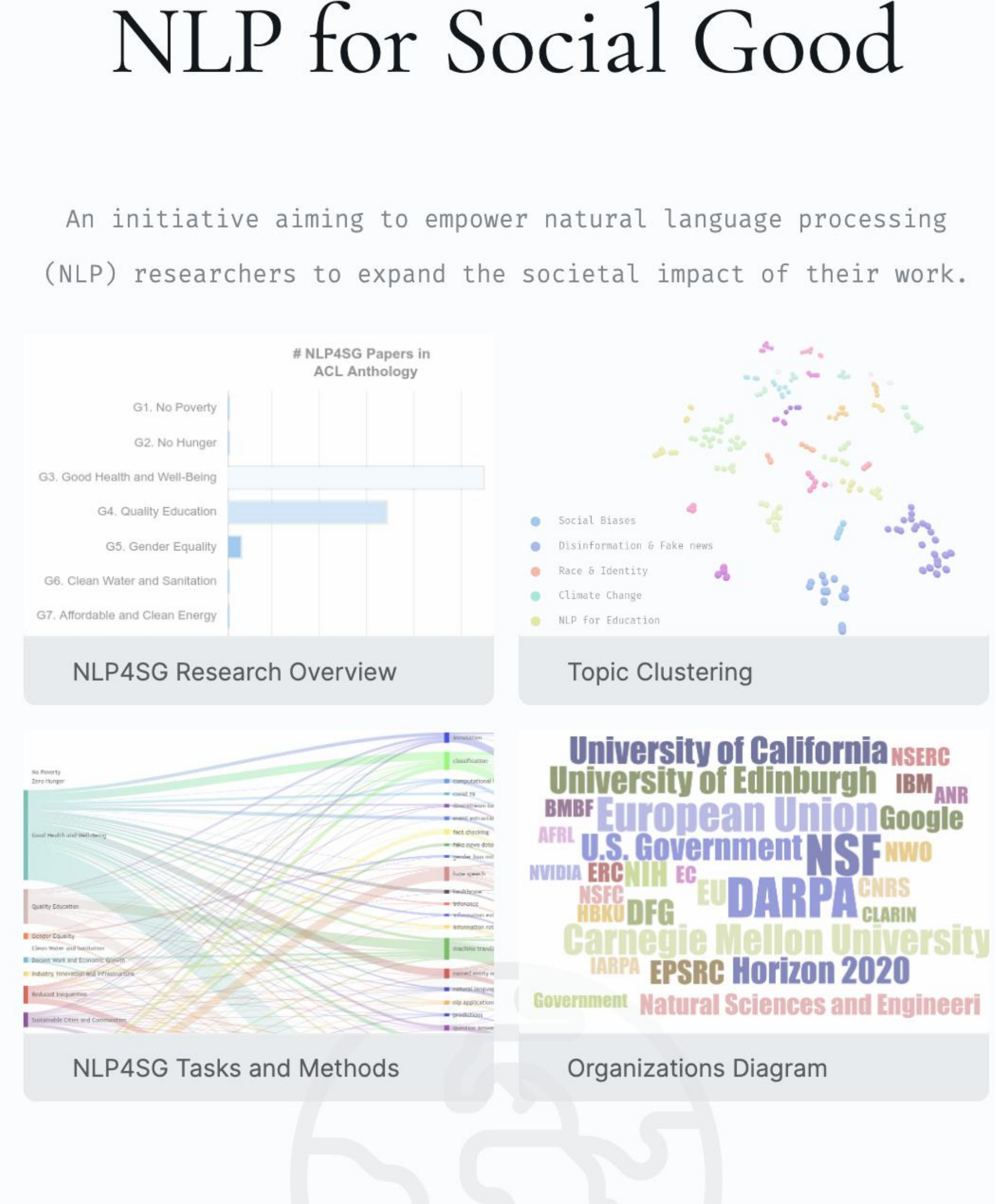}
    \fi
    \caption{The homepage of our NLP4SG website.}
    \label{fig:website_homepage}
\end{figure}

Using our NLP4SG \ourmodel system, we analyze all the 76K NLP papers from ACL Anthology to answer our driving questions \cref{item:q1}, \cref{item:q2}, and \cref{item:q3}.
Additionally, we also build a demo website at 
\ifarxiv
\href{https://nlp4sg.vercel.app}{\texttt{nlp4sg.vercel.app}}
\else
[Anonymous Link]
\fi
to make our analysis results more accessible to all parties interested in NLP4SG. 
Our website starts with a homepage in 
\cref{fig:website_homepage}, whose visualization plots we will refer to for the related analysis in this section. For a full overview of various demos on the website, see \cref{appd:screenshots}.

\begin{figure}[ht]
    \centering
    \includegraphics[width=\columnwidth]{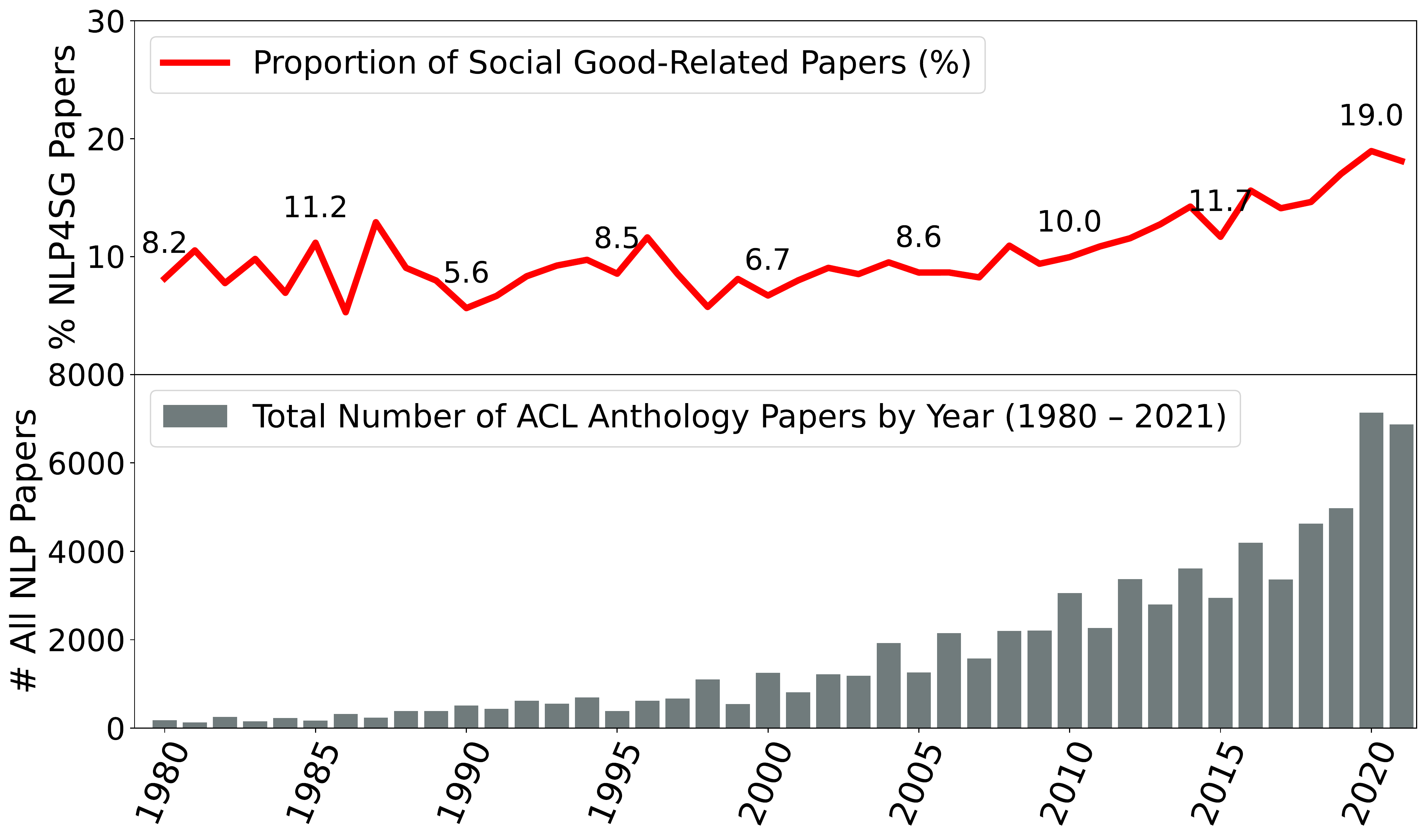}
    \caption{Percentage of social good-related papers (top) and the total number of papers (bottom) in ACL Anthology each year from 1980. }
    \label{fig:dist_papers}
\end{figure}

\paragraph{Q1. How Many NLP4SG Papers Are There?}
We apply our NLP4SG binary classifier to the entire ACL Anthology with 76K papers. 
Overall, there are around 13.07\% social good-related papers.
For a more detailed overview, we plot the proportion of social good-related papers among all NLP papers every year from 1980 in \cref{fig:dist_papers}. We can see a clear rising trend of NLP4SG in recent years. Notably, the proportion of NLP4SG papers increased to 19.0\% in 2020 from the initial percentage of 8.2\% in 1980. It is important to note that while these percentages provide us with directional insight into the trend, their accuracy is contingent on the precision of the model used for analysis.

\paragraph{Q2. What Aspects of Social Good Do the Papers Address?}\label{sec:insight_sdg}
Next, we are interested in the question of what social good aspects are addressed by NLP research.

\begin{figure}[ht]
    \centering
    \includegraphics[width=\columnwidth]{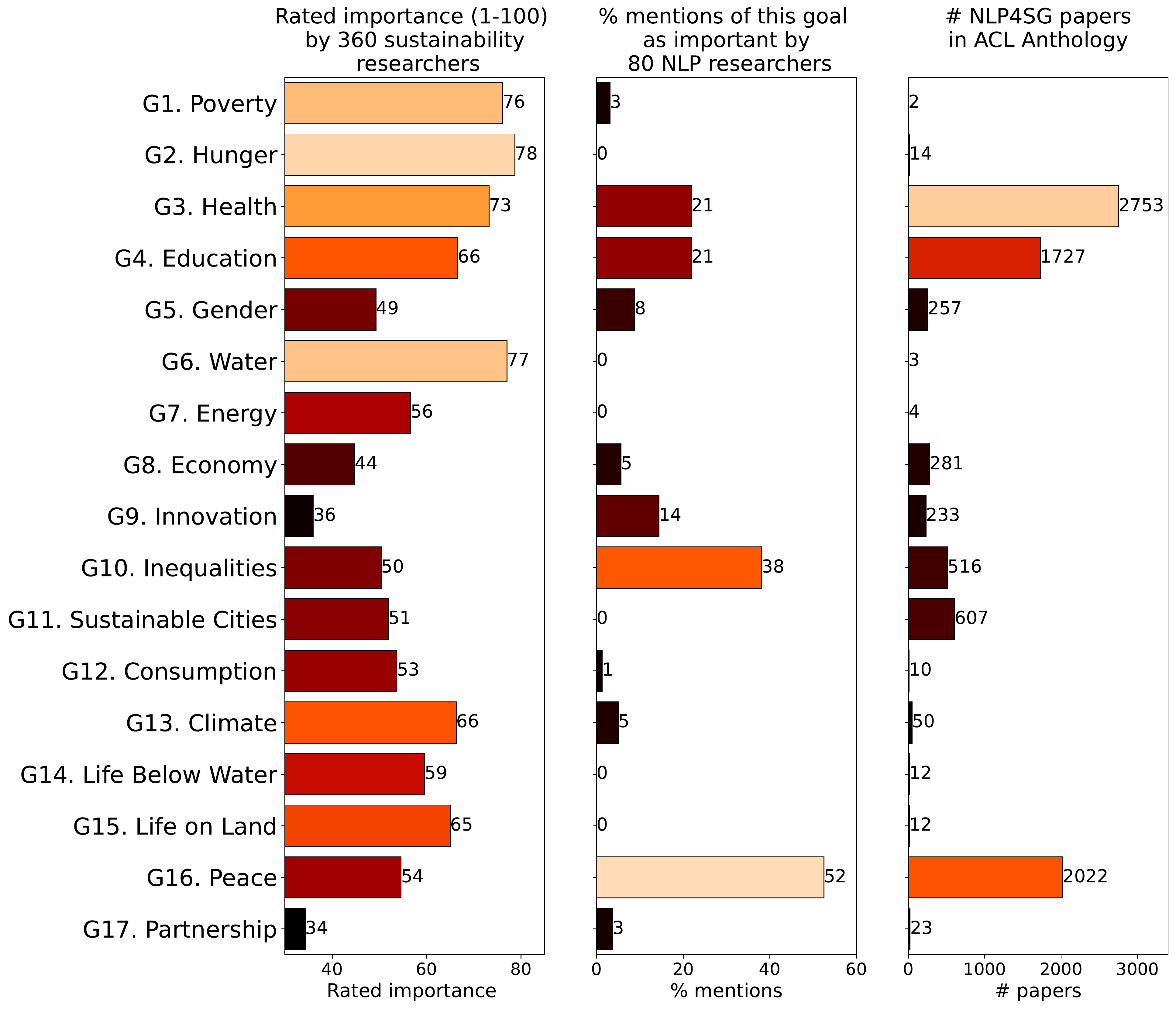}
    \caption{SDGs importance vs actual efforts. We use darker red for shorter bars, so the overlooked goals (especially on the right chart) can draw more attention.}
    \label{fig:comparison}
\end{figure}

In \cref{fig:comparison}, we show three bar charts side by side. The left one shows the importance scores of all the 17 SDGs according to \citet{yang2020prioritizing}'s worldwide survey of 360 sustainability researchers, where they incorporate all expert ratings into an overall priority score. 
The chart in the middle is compiled from
our survey on ``How current NLP researchers think about NLP4SG''
\ifarxiv
at ACL 2021
\fi
of 80 NLP researchers, where we ask the subjects ``What social good applications do you think the NLP community should consider?'' and then map the answers (e.g., NLP for education, and combating misinformation) to corresponding SDGs. 
The survey details are in \cref{appd:survey}.
Finally, we apply our UN SDG goal classifier to all the ACL Anthology papers, and visualize the distribution of the 17 goals into the right panel of \cref{fig:comparison}. 
Our demo website\ifarxiv \footnote{\url{https://nlp4sg.vercel.app/sdg}}
\fi 
~contains a version of this tri-view comparison in \cref{appd:website_sdg}, where the user can specify whether they want to look at the overall trend, or the trend in any specific year.

From the overall comparison in \cref{fig:comparison}, we can see a strong mismatch between the importance of the goals (left) and the actual size of research efforts (right). For example, NLP4SG papers mostly address health, peace, and education, which are admittedly important goals. However, they hardly address other important goals such as poverty and climate, which are pressing social issues. 
This might be because it may be more challenging to address these topics, e.g., lack of available datasets and expertise, trends dominating researchers' attention, funding climate (e.g., healthcare research being better funded), ease of publishing papers, the popularity of the field which may help gain citations more easily, the existence of good benchmark datasets, and so on.
Note that we intentionally avoid getting into technological solutionism, and a more accurate framing should be trying to make the best use of the space where NLP can help (\cref{appd:discuss_alignment}), and the first two bar charts are all references to help gain better insights into this.

\myparagraph{Q3. What Technical Solutions Are There?}
Diving deeper into our analysis, 
we study
how existing research formulates tasks for each social good goal, and what methods they tend to adopt.

We visualize the distribution of tasks and methods in a Sankey diagram at the bottom of \cref{fig:overview}, and add the connections to the SDGs. Briefly, the Sankey diagram takes as input the SDG-task-method information identified from each of the 9K NLP4SG papers from ACL Anthology.
Hence, the diagram forms a comprehensive overview of what goals are addressed by what kind of task formulations, and then solved by what kind of models. Thus, researchers can quickly gain insights into NLP4SG and get an overview of the technical landscape in one place, and then spot opportunities where a better solution is yet to be proposed, e.g., new task proposals for an under-addressed goal such as poverty, or methodological upgrades for existing tasks so that the field can benefit from the latest advancement of NLP models.

The most common tasks in NLP4SG papers are: machine translation, which is present in 6.30\% of the papers; classification in 4.06\%; toxicity detection in 3.32\%; named entity recognition in 2.19\%; and sentiment analysis in 1.51\%. The most common methods are: LLMs present in 12.18\% of the NLP4SG papers, classifiers in 2.98\%, deep neural networks in 2.51\%, word embeddings in 1.80\%, and SVMs in 1.62\%.

\begin{figure}[ht]
    \centering
    \includegraphics[width=\columnwidth]{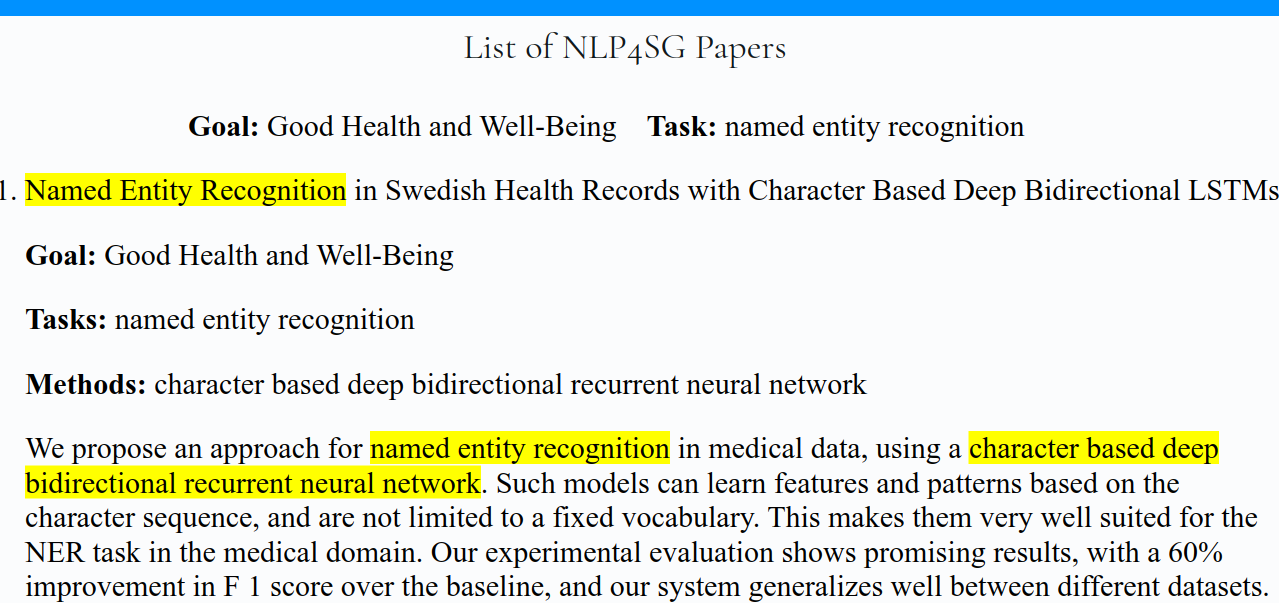}
    \caption{An example paper in the paper list after clicking on ``Health'' and ``Named Entity Recognition.''}
    \label{fig:papers_list}
\end{figure}

On our Sankey diagram demo page,\ifarxiv
\footnote{\url{https://nlp4sg.vercel.app/sankey}}
\fi
~we also enable clicks into each goal, task, or method to see the corresponding list of papers.
We show the titles and abstracts of the papers and highlight text spans that contain the tasks and methods, as illustrated in \cref{fig:papers_list}. This function is helpful to facilitate NLP4SG literature review, and saves time for future researchers. 

\section{Related Work}
\myparagraph{AI for Social Good.} 
Recent years have seen a rising effort to apply AI technologies to various aspects of social good \cite[e.g.,][]{chen2020artificial,yu2018artificial}. 
To enhance the community which cares about AI for social good \cite{tomavsev2020ai,jin-etal-2021-good,hager2019artificial},
it is crucial to gain insights about the progress overview.
However, existing efforts mostly rely on time-consuming manual compilation, from surveying hundreds of papers \cite{cowls2021definition,jin-etal-2021-good}, conducting limited case studies \cite{floridi2020design,hager2019artificial}, to pairing existing papers on tasks and datasets with SDGs \cite{yeh2021sustainbench}. Some work uses keyword matching to extract social good-related papers \cite{shi2020artificial,fortuna-etal-2021-cartography}, but our work is the first to build a comprehensive paper analysis system, and we are the only work designing a suit of various tasks, annotating a dataset, and using state-of-the-art NLP models to support such analysis.

\myparagraph{NLP on Scientific Literature.} is an active research area on building NLP models for analysing scientific text. Example research directions include
dataset creation \cite{ammar-etal-2018-construction,lo-etal-2020-s2orc,wang2020cord}; modeling for tasks such as information extraction \cite{jain-etal-2020-scirex,luan-etal-2018-multi,zhong-chen-2021-frustratingly,hope-etal-2021-extracting}, 
summarization \cite{cachola-etal-2020-tldr}, question answering \cite{dasigi-etal-2021-dataset}, and general language modeling \cite{beltagy2019scibert,taylor2022galactica}. There are notable visualization platforms e.g. for COVID-19 research papers, such as search engines to browse papers \cite{bhatia2020aws}, challenges and directions \citep{lahav2022search}, and associations between research groups and topics \cite{hope-etal-2020-scisight}. Our work has a unique focus on identifying NLP4SG papers, aligning them with SDGs, and visualizing the research landscape, and it is the first visualization platform for NLP4SG based on NLP papers.

\section{Conclusion}
In this work, we make possible the reporting of the state of NLP4SG, with the help of several novel tasks specific to NLP4SG papers, the construction of a richly annotated \ourdata dataset, and the development  of a \ourmodel system based on state-of-the-art LLMs.
We use this system to provide researchers insights into the landscape of NLP4SG research,
paving the way for a more evidence-driven pursuit of NLP4SG that goes beyond good intentions.
To make this information more accessible, we also create a website to visualize our analysis.
\ifarxiv
\else All the website visualizations are available in the supplementary materials, and will be released with the camera ready.
\fi

\section*{Limitations}
One limitation of this work is that the models are not yet perfect, so the insights on the entire ACL Anthology are bottlenecked by the best model performance. To make up for this, we also provide in \cref{appd:gold_demo} an additional analysis with only the gold, annotated data, which are accurate, but smaller and less representative of the entire landscape of NLP4SG.
Since our system is modular, and we can plug in better models for each task, we welcome future work to further improve the current models. One example improvement direction is to model more inputs than the title and abstract, for instance by taking into consideration the full text and meta information such as the venue and publication year of the paper.

Our sampling technique is designed to ensure the representativeness of our sample within the larger population of academic papers in our dataset, in line with our primary objective to reflect the distribution of each SDG. We acknowledge that there may be variations in the distribution of papers over the years, particularly in earlier years with limited number of papers. In future work, a more detailed analysis could be conducted to track the distribution of NLP4SG papers over time, providing a more nuanced understanding of how these papers have evolved and complementing our broader findings regarding SDG representation.

Another limitation is that the definition of social good might experience a gradual shift when society evolves its ethical standards. There is no fixed, definitive answer, so we welcome community discussions. This paper is a snapshot of existing thinking frameworks \cite{cowls2021definition,jin-etal-2021-good} and our understanding by 2023. Updating the definition of social good is key to mitigating the risks of overlooking important societal problems. In the long run, we suggest community-wide discussions and iterations to update the definition of social good over time.

Finally, beyond the three tasks we propose for NLP4SG papers, there could be potentially other tasks, such as generating task formulations for under-addressed goals, suggesting new task-method combinations, and so on. For example, some NLP tasks that could address issues related to poverty could include sentiment analysis to understand public perceptions of poverty and related issues, information extraction from government documents to identify and track funding for poverty reduction programs, and machine translation to make information about poverty reduction programs and resources available to non-English speakers.
Moreover, currently our work is limited to the knowledge within NLP, so it will be better if future work can collect data about the downstream real-world impact of each research trace, which can provide better feedback for the community.

\section*{Ethical Considerations}
This work aims to provide insights about the state of NLP4SG research. It helps to increase the awareness of NLP researchers about applying NLP for positive social applications. In addition, the data we use is based on the open-sourced ACL Anthology dataset, so there are no user privacy concerns. To make this line of work more inclusive and comprehensive, we suggest a regular update of the social good identifier to keep up with the ethical standards of the community.

\ifarxiv
\section*{Author Contributions}
The idea and mission of the paper were mainly devised by \textit{Zhijing} and \textit{Rada} as an important step in the NLP for Social Good initiative.

\textit{Fernando} put tremendous efforts into running all the models, building the demo website, and various analyses.
\textit{Zhijing} guided the work, and supervised the entire implementation process. 
\textit{Fernando} and \textit{Zhijing} equally contributed to the dataset collection.

\textit{Mrinmaya} supervised Fernando's Master thesis, from which this work extended.
\textit{Mrinmaya} and \textit{Bernhard} gave high-level feedbacks of the project.
\textit{Tom} gave insightful suggestions about how to design the paper and the dataset, as well as the common NLP models to use for scientific text.
\textit{Zhijing}, \textit{Fernando}, \textit{Mrinmaya} and \textit{Rada} led the paper writing, and all authors contributed to improving the writing.
\fi

\ifarxiv
\section*{Acknowledgments}

We thank Niklas Stoehr for his feedbacks on our earlier work \cite{jin-etal-2021-good} and his suggestion that we could further automate that framework. We appreciate 
Dan Lahav for his insightful discussions to improve the orientations of this paper and its role in NLP4SG initative. 
We acknowledge Jad Beydoun for helping with the dataset collection process, and Flavio Schneider for deploying the basic structure of the website based on which we built the visualizations. 
We thank Ariel Holtzman for the brilliant brainstorming of the paper title,
Laura Biester for insightful suggestions when proofreading the paper, and LIT labmates Ashkan Kazemi and Artem Abzaliev for suggestions to improve the messages that this paper convey.

This material is based in part upon works supported by the Swiss National Science Foundation (Project No. 197155); a Responsible AI grant by the Haslerstiftung; an ETH Grant (ETH-19 21-1); the German Federal Ministry of Education and Research (BMBF): Tübingen AI Center, FKZ: 01IS18039B; the Machine Learning Cluster of Excellence, EXC number 2064/1 – Project number 390727645; and by the John Templeton Foundation (grant \#61156).
Zhijing Jin is supported by PhD fellowships from the Future of Life Institute and Open Philanthropy. 
\fi

\bibliography{sec/refs_acl,sec/refs_this_paper,sec/refs_causality,sec/refs_zhijing,sec/refs_ai_safety,sec/refs_cogsci,sec/refs_semantic_scholar,sec/refs_nlp4sg}
\bibliographystyle{acl_natbib}
\clearpage
\appendix

\section{Overview of the Demo Website }\label{appd:screenshots}
We create a demo website at 
\ifarxiv
\url{https://nlp4sg.vercel.app/}.
\else
[Anonymous Link].
\fi
Our website starts with a homepage in 
\cref{fig:website_homepage}, which provides access to all the analysis visualization plots made in Javascript. We introduce each visualization function in detail below.

\subsection{Visualization 1: SDG Alignment}\label{appd:website_sdg}
\begin{figure}[ht]
    \centering
    \ifshowfig
    \includegraphics[width=\columnwidth]{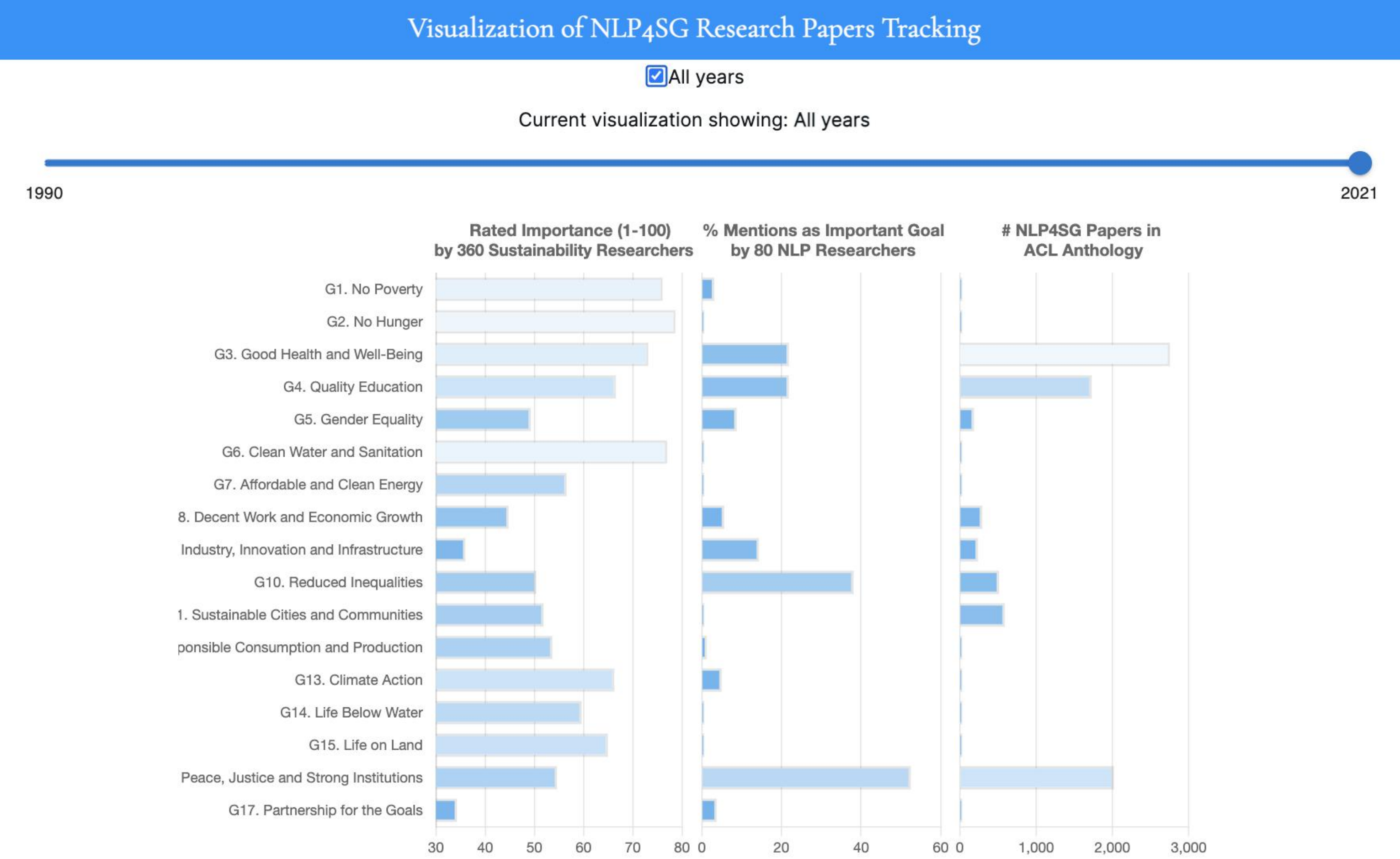}
    \fi
    \caption{Visualization of the importance of/efforts on the SDGs by the sustainability researchers, NLP researchers, and actual ACL Anthology papers.}
    \label{fig:website_sdg}
\end{figure}

We first provide a visualization in \cref{fig:website_sdg} of the SDGs (\cref{sec:insight_sdg}) from three aspects: the importance scores of SDGs collected in a survey on sustainability researchers \cite{yang2020prioritizing}, NLP researchers' opinions, and the number of ACL Anthology papers addressing the goals. In addition to the static plot in \cref{fig:comparison}, we enable a scroll bar on the top to select the view of papers only in a certain year at \ifarxiv
\url{https://nlp4sg.vercel.app/sdg}.
\else
[Anonymous Link].
\fi

\subsection{Visualization 2: Sankey Diagram of the Research Landscape} The Sankey diagram of the research landscape is shown in the bottom of \cref{fig:overview} or it can be directed accessed at \ifarxiv
\url{https://nlp4sg.vercel.app/sankey}.
\else
[Anonymous Link].
\fi
We include in \cref{appd:website_implementation} the details of the scientific term normalization to clean the data for the Sankey diagram.

\subsection{Visualization 3: Paper Browser} \label{appd:paper_browser} 
Corresponding to the Sankey diagram, we also enable clicks into each goal, task, or method to see the corresponding list of papers, an example of which is shown in \cref{fig:papers_list}\ifarxiv
~(\url{https://nlp4sg.vercel.app/sankey})\fi.
We show the titles and abstracts of the papers and highlight text spans that contain the tasks and methods. This function is helpful to facilitate NLP4SG literature review, and saves time for future researchers.

\subsection{Visualization 4: Supporting Organizations} 
\begin{figure}[ht]
    \centering
    \ifshowfig
    \includegraphics[width=0.6\columnwidth]{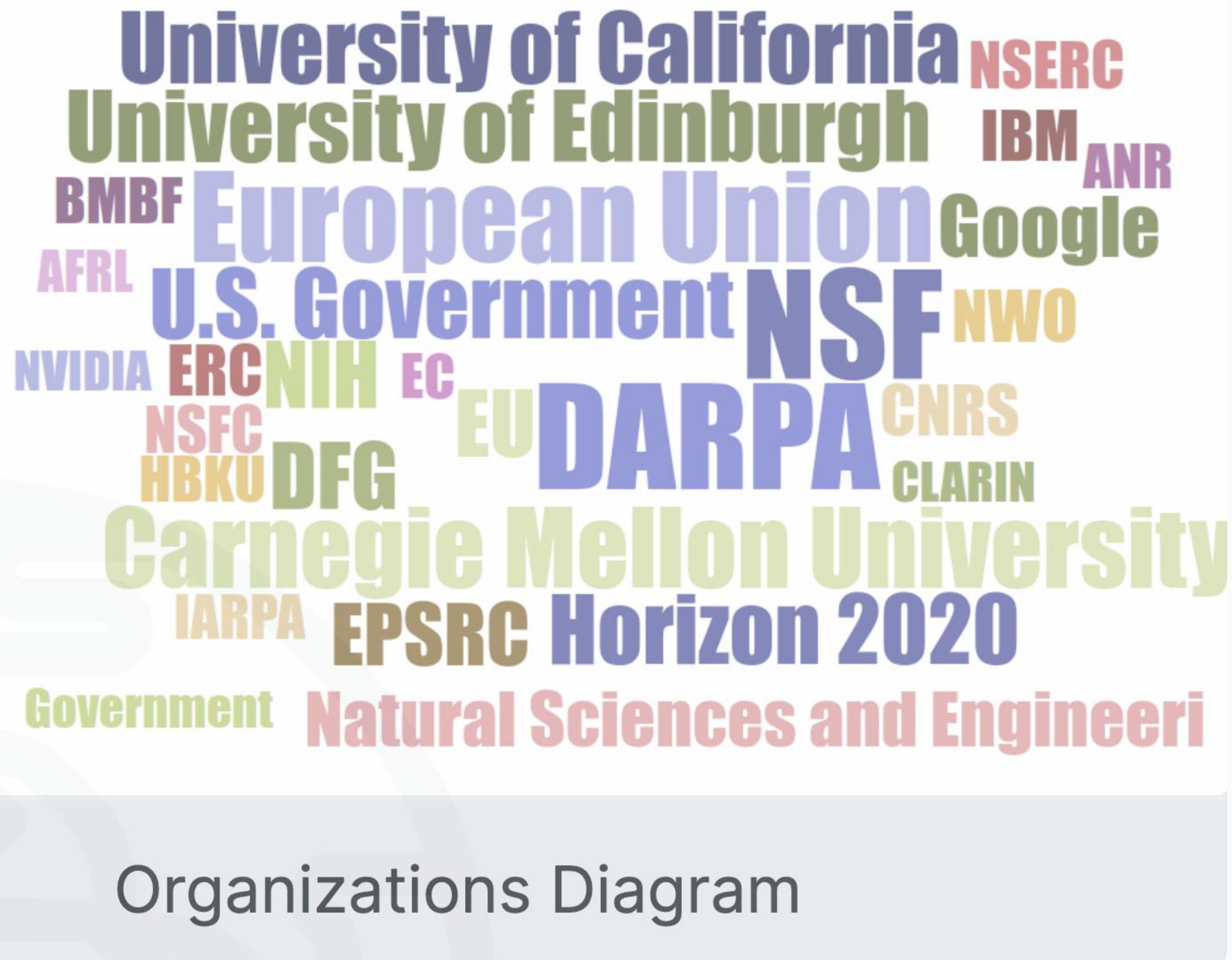}
    \fi
    \caption{Word cloud of the organizations mentioned in the NLP4SG papers.}
    \label{fig:website_org_wordcloud}
\end{figure}

\begin{figure}[ht]
    \centering
    \ifshowfig
    \includegraphics[width=\columnwidth]{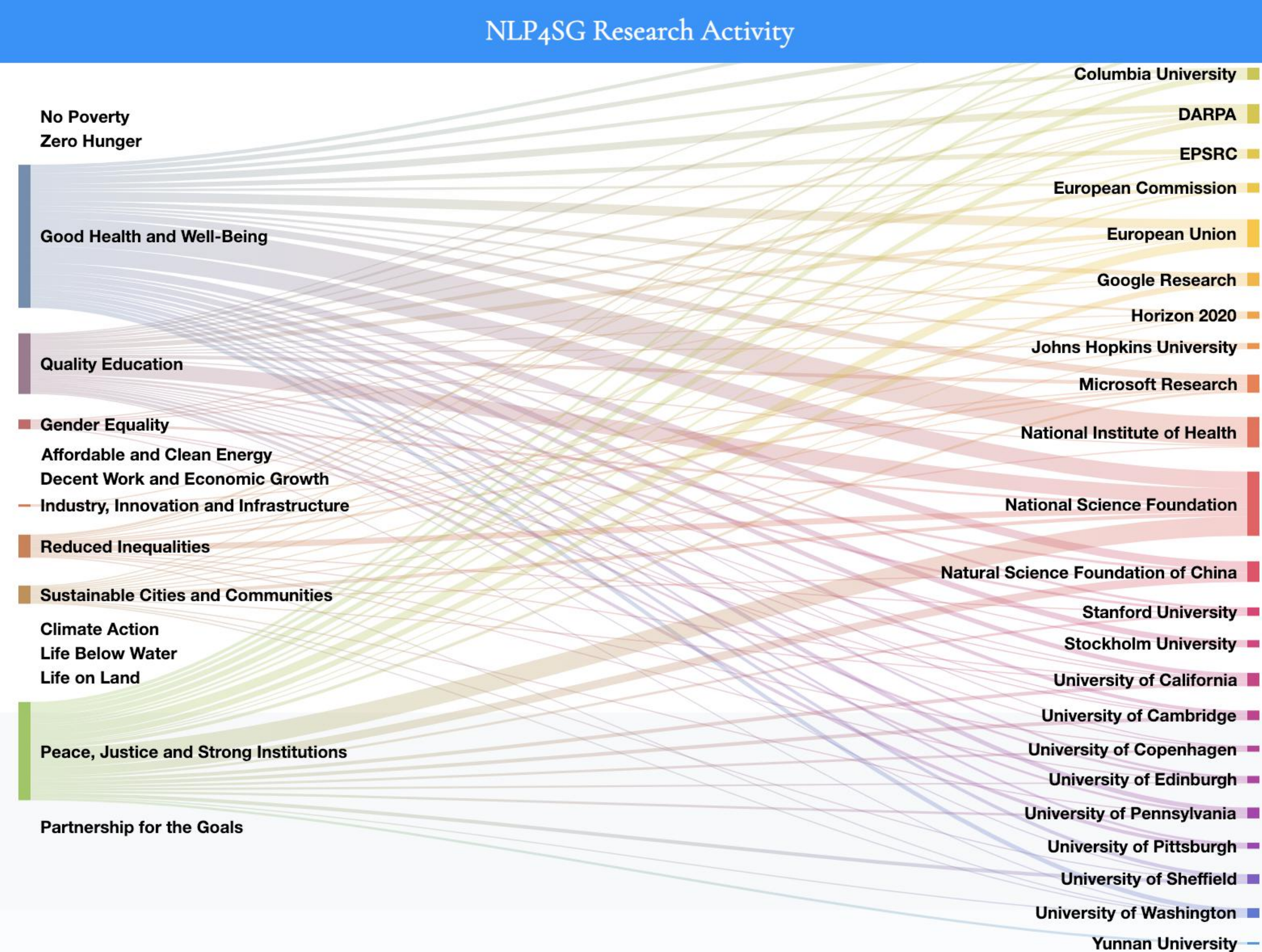}
    \fi
    \caption{Sankey diagram of the organizations supporting the NLP4SG research work on each SDG.}
    \label{fig:website_org_sankey}
\end{figure}

To understand the support structure behind the pursuit of the SDGs, we collect the organizations supporting each research paper. We ran named entity recognition models by the Stanza Python package \cite{qi2020stanza} over the Acknowledgment section. In case no organizations can be identified, then we use hand-crafted rules such as taking the main organization name of the authors if it is not a research institute, as these organizations could be self-supported.
We visualize the distribution of organizations in the forms of a word cloud in \cref{fig:website_org_wordcloud} and a Sankey diagram corresponding to each SDG in \cref{fig:website_org_sankey}.

\subsection{Visualization 5: Paper Cluster Visualization} 
\begin{figure}[ht]
    \centering
    \ifshowfig
    \includegraphics[width=\columnwidth]{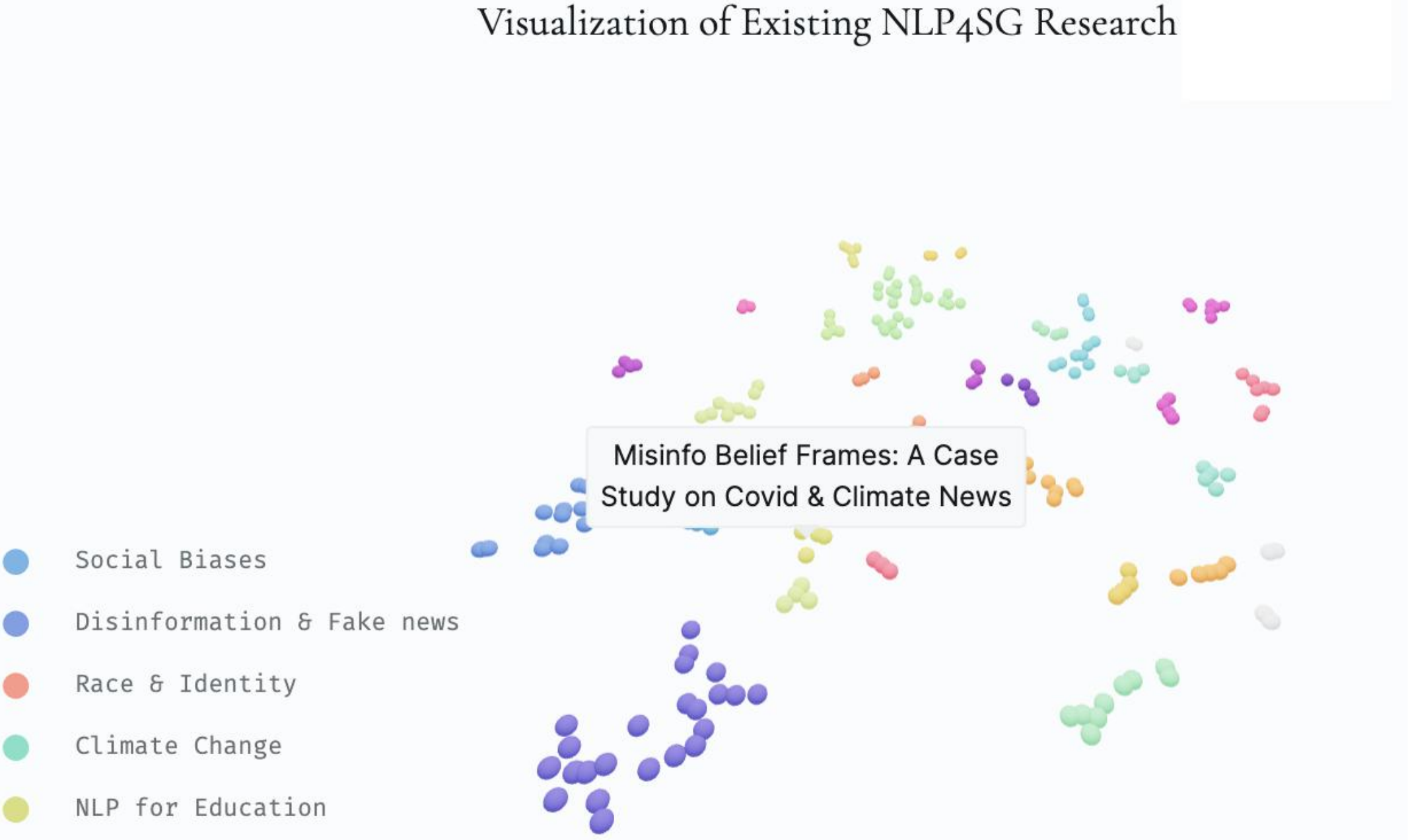}
    \fi
    \caption{Clustering of the NLP4SG papers.}
    \label{fig:website_topic}
\end{figure}

We also make a bottom-up plot of all the papers at a glance at \ifarxiv
\url{https://nlp4sg.vercel.app/papers}.
\else
[Anonymous Link].
\fi 
Specifically, we use the text embeddings of the papers by sentence transformer  \cite{reimers-gurevych-2019-sentence},\footnote{We use the \texttt{all-mpnet-base-v2} model of huggingface at \url{https://huggingface.co/sentence-transformers/all-mpnet-base-v2}}
project them using t-SNE \cite{rauber2016visualizing}, and cluster by the HDBSCAN clustering algorithm \cite{campello2013densitybased}.
A screenshot of the paper clustering visualization page is in \cref{fig:website_topic}.

\section{Data Collection Details}\label{dataset_details}

\subsection{Data Source}\label{appd:data_source}
We identify ACL Anthology as a quality source of data, since it contains a large set of peer-reviewed NLP papers. There are other potential sources of peer-reviewed NLP papers, such as papers submitted to the NLP track at other venues such as NeurIPS and ICLR; as well as non-peer-reviewed papers on arXiv under the category cs.CL. However, for the former, it is hard to algorithmically identify NLP papers with high precision and recall, not to mention the different copyright regulations; for the latter, it is hard to control the paper quality if we open to non-peer-reviewed data. After balancing all the concerns and also thinking of our target audience, the NLP community, we think ACL Anthology, which is peer-reviewed and maintained by the NLP community, is a good data source for our work.

\subsection{Preprocessing}\label{appd:data_preprocessing}
We conduct the following preprocessing on the ACL Anthology data. 
To identify the paper entries in the entire anthology, we remove those that are workshop proceedings, invited talks, and book reviews, by a set of hand-crafted rules, such as identifying the title starting with ``Book Reviews: \dots'' or ending with ``\dots Workshop.'' 
To clean the titles, we remove the special symbols ``\{'' and ``\}'' in titles such as ``\{N\}o\{P\}ropaganda at \{S\}em\{E\}val-2020 Task 11: A Borrowed Approach to Sequence Tagging and Text Classification.''

Moreover, not all entries on the anthology provide a non-empty abstract, so for papers with missing abstracts, we parse their PDFs using the PDF-to-JSON converter \texttt{doc2json} Python package\footnote{\url{https://github.com/allenai/s2orc-doc2json}} to extract the abstracts.

\subsection{Annotation Scheme}\label{appd:annotation_guide} 
We introduce the annotation scheme for the three tasks below.

\subsubsection{Annotating Task 1: NLP4SG Classification}

\begin{figure}[ht]
    \centering
    \includegraphics[width=\columnwidth]{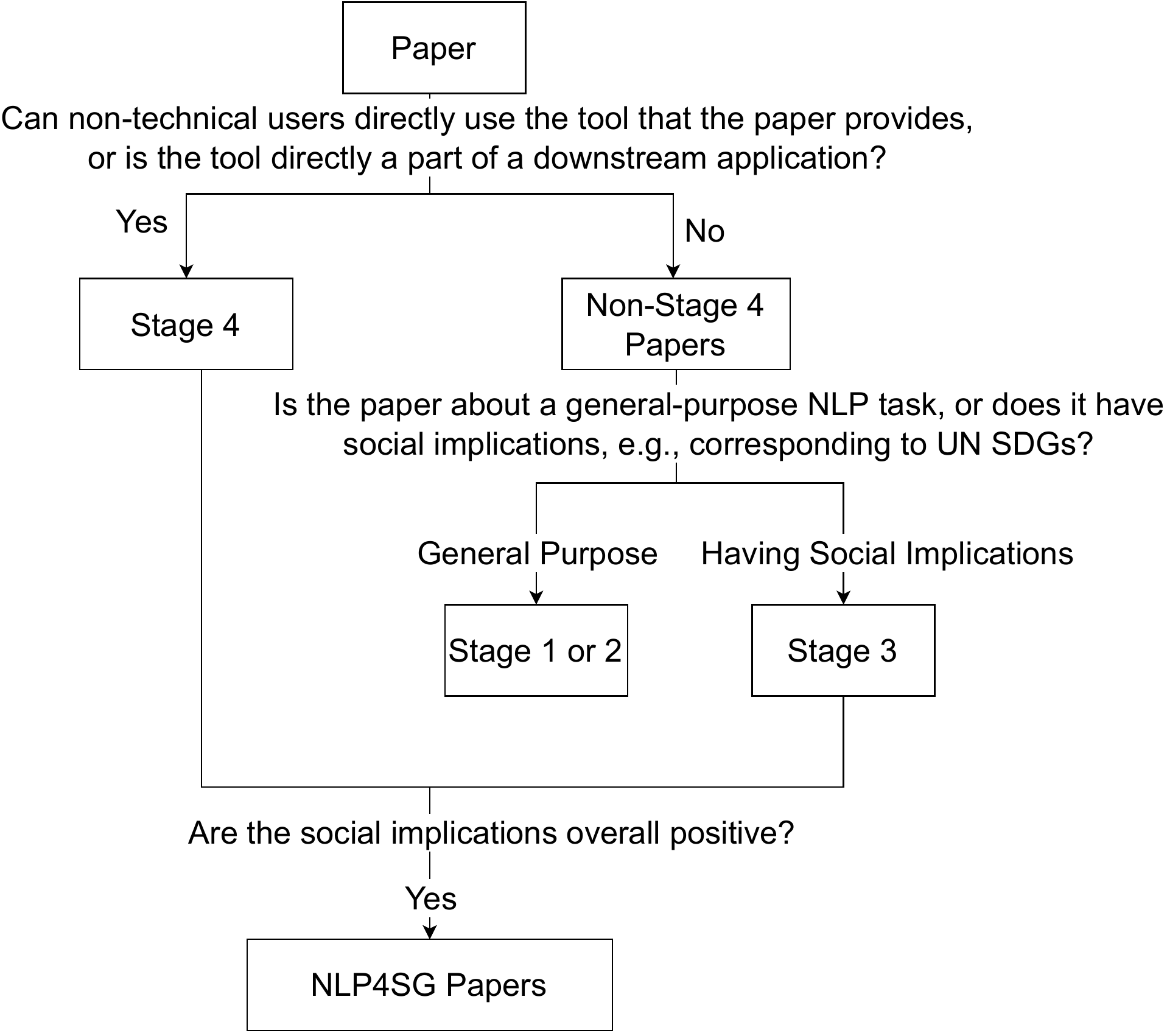}
    \caption{A decision flowchart used for the annotation of Task 1, NLP4SG binary classification.}
    \label{fig:rules_sg}
\end{figure}

Inspired by previous frameworks defining NLP4SG 
\cite{jin-etal-2021-good} and AI for social good \cite{cowls2021definition}, we design a decision flowchart to identify whether an NLP paper addresses a social good-related problem in \cref{fig:rules_sg}.

Specifically, there are two important decision criteria. One is about the relevance to social good topics, e.g., the 17 UN SDGs. We also include social problems in the digital era, such as online toxicity, fake news, and privacy, into their corresponding SDG, i.e., (Goal 16) Peace, justice and strong institutions.
The second criterion is about how directly the paper addresses the social problem. For example, general linguistic tools such as syntax parsing could be an important technology, but not carrying direct social implications. We adopt the categorization of the stage of a technology in \citet{jin-etal-2021-good}, and only count those with a clear implication for downstream applications (i.e., Stage 3 and 4).

To get the idea through to the annotators, we provide both the formal guideline above, and also concrete examples as follows.

\textit{What is included:}
(Example set 1) Directly related to the high-level definition of SDGs, mentioning, e.g., healthcare; mental health (psycho-counseling, hope speech); education; facilitating efficient scientific research (which belongs to Goal 9 Industry, innovation and infrastructure); helping employment (job matching, training job skills); helping collaboration among decision-makers. (Example set 2) Related to the fine-grained subcategories of SDGs, e.g., encouraging civic engagement, and enabling social problem tracking for the goal of (Goal 16) Peace, justice and strong institutions.
(Example set 3) Social problems in the digital era: e.g., online toxicity, misinformation, privacy protection, and deception detection.

\textit{What is excluded:}
(Example set 1) General-purpose, coarse-grained NLP tasks: machine translation, language modeling, summarization, sentiment analysis, etc. 
(Example set 2) General-purpose, fine-grained NLP tasks:
news classification; humor detection; technologies to increasing productivity, e.g., email classification, report generation, meeting note compilation (because they are application-agnostic which could be used for both good and bad purposes, and also a bit too general); textbook-related QA but using it as a benchmark to improve the general modeling capabilities; tasks whose data is socially relevant, but the task is neutral (e.g., POS tagging for parliament speech); NLP to help other neutral disciplines, e.g., chemistry; tasks a bit too indirectly related to SDGs, e.g., parsing historical language document, or cultural heritage-related tasks, low-resource MT, which bridges resources from one community to another, but is a bit too indirect, and also depends case by case on the actual language community, plus there is a tradeoff between efficiency and equality.
(Example set 3) 
Tasks with controversial nature or unknown effect (varying a lot by how people use them in the future): e.g., news comment generation; financialNLP, which could be used in either way to help the economy, or perturb the market for private profits; simulated NLP tools for the battlefield; user-level demographic prediction.

As mentioned in the Limitations and Ethical Implications sections, the definition of social good might experience a gradual shift when society evolves its ethical standards. There is no fixed, definitive answer, so we start with relatively inclusive criteria here, and welcome community discussions and iterations to improve the definitions. This paper is a snapshot of existing thinking frameworks \cite{cowls2021definition,jin-etal-2021-good} and our understanding by 2023. In the long run, we suggest community-wide discussions and iterations to update the definition of social good from time to time.

\subsubsection{Annotating Task 2: SDG Classification}\label{appd:sdg_def}

We let the annotators read the official \textit{descriptions} and \textit{indicators} of the 17 SDGs on the UN website such as \url{https://sdgs.un.org/goals/goal1}.
For each paper identified as NLP4SG in Task 1, we ask the annotators to label all relevant SDGs.

We present examples of NLP4SG paper titles corresponding to each SDG below.
\myparagraph{G2. Hunger}
\begin{itemize}
\item A Gold Standard for CLIR evaluation in the Organic Agriculture Domain
\item CRITTER: a translation system for agricultural market reports
\end{itemize}
\myparagraph{G3. Health}
\begin{itemize}
\item A Treebank for the Healthcare Domain
\item Automatic Analysis of Patient History Episodes in Bulgarian Hospital Discharge Letters
\end{itemize}
\myparagraph{G4. Education}
\begin{itemize}
\item An {MT} learning environment for computational linguistics students
\item Salinlahi {III}: An Intelligent Tutoring System for {F}ilipino Heritage Language Learners
\end{itemize}
\myparagraph{G5. Gender}
\begin{itemize}
\item An Annotated Corpus for Sexism Detection in {F}rench Tweets
\item Mitigating Gender Bias in Machine Translation with Target Gender Annotations
\end{itemize}
\myparagraph{G6. Water}
\begin{itemize}
\item A conceptual ontology in the water domain of knowledge to bridge the lexical semantics of stratified discursive strata
\end{itemize}
\myparagraph{G8. Economy}
\begin{itemize}
\item Multilingual Generation and Summarization of Job Adverts: the {TREE} Project
\item Situational Language Training for Hotel Receptionists
\end{itemize}
\myparagraph{G9. Innovation}
\begin{itemize}
\item An Annotated Corpus for Machine Reading of Instructions in Wet Lab Protocols
\item Retrieval of Research-level Mathematical Information Needs: A Test Collection and Technical Terminology Experiment
\end{itemize}
\myparagraph{G10. Inequalities}
\begin{itemize}
\item Analyzing Stereotypes in Generative Text Inference Tasks
\item Recognition of Static Features in Sign Language Using Key-Points
\end{itemize}
\myparagraph{G11. Sustainable Cities}
\begin{itemize}
\item FloDusTA: Saudi Tweets Dataset for Flood, Dust Storm, and Traffic Accident Events
\item Trouble on the Road: Finding Reasons for Commuter Stress from Tweets
\end{itemize}
\myparagraph{G12. Consumption}
\begin{itemize}
\item Multiple Teacher Distillation for Robust and Greener Models
\end{itemize}
\myparagraph{G13. Climate}
\begin{itemize}
\item CLIMATE-FEVER: A Dataset for Verification of Real-World Climate Claims
\item Tackling Climate Change with Machine Learning
\end{itemize}
\myparagraph{G14. Life Below Water}
\begin{itemize}
\item Marine Variable Linker: Exploring Relations between Changing Variables in Marine Science Literature
\item Literature-based discovery for Oceanographic climate science
\end{itemize}
\myparagraph{G16. Peace}
\begin{itemize}
\item On Unifying Misinformation Detection
\item Fully Connected Neural Network with Advance Preprocessor to Identify Aggression over Facebook and Twitter
\end{itemize}
Since Goal 16 is a big category including tasks such as toxicity detection, and fighting misinformation, we provide an additional label when a paper within Peace addresses more fine-grained tasks. We collect statistics for the main subcategories: 
44 samples (23.28\%) for toxicity detection, 44 samples (23.28\%) for fighting misinformation, 11 samples (5.82\%) for privacy protection, and 6 samples (3.17\%) for deception detection.

\myparagraph{G17. Partnership}
\begin{itemize}
\item MEDAR: Collaboration between European and Mediterranean Arabic Partners to Support the Development of Language Technology for Arabic
\item The Telling Tail: Signals of Success in Electronic Negotiation Texts
\end{itemize}

\subsubsection{Annotating Additional Data for Task 2: Upsampling}\label{appd:upsampling}

To enrich the data for Task 2, we upsample the low-occurrence classes. We denote all the classes with less than 5\% proportion in our dataset as the low-occurrence classes.
We first automatically extract candidate papers that might correspond to the low-occurrence classes, and then manually annotate their SDG classes. 
To automatically identify a candidate pool of papers regarding the low-occurrence classes, we first run the best model of Task 1 on the entire ACL Anthology to extract 9K NLP4SG papers, and among them, we use our curated keywords corresponding to the SDGs to extract candidate papers.

In \cref{tab:upsample_details}, we can see that by the automatic matching algorithm, we get 261 candidate papers in total, including 1 candidate paper for poverty, 25 for hunger, 67 for gender, ..., and 40 for partnership. And after manually annotating these 261 papers, we obtain 4 confirmed papers for hunger, 67 for gender, 1 for water, 44 for economy, ..., and 5 for partnership.
We use this upsampled data as additional data for Task 2, resulting in 759 total NLP4SG papers for Task 2.

\begin{table}[ht]
    \centering\small
    \setlength\tabcolsep{2.7pt}
    \resizebox{\columnwidth}{!}
    {%
    \begin{tabular}{lcc}
\toprule
\textbf{UN SDG} & \# Matched Papers & \# Papers after Annotation \\ \hline
G1. Poverty & 1  &  0\\
G2. Hunger & 25  &  4\\
G3. Health & High Occurrence &  0 \\
G4. Education & High Occurrence & 0 \\
G5. Gender & 67   &  67\\
G6. Water & 3  & 1 \\
G7. Energy & 2   &  0\\
G8. Economy & 67  & 44 \\
G9. Innovation & High Occurrence  & 15 \\
G10. Inequalities & High Occurrence & 1   \\
G11. Sustainable Cities & 33 & 11 \\
G12. Consumption & 0  &  0\\
G13. Climate & 21  &  19\\
G14. Life Below Water & 2  & 2 \\
G15. Life on Land & 0   &  0\\
G16. Peace & High Occurrence &  12 \\
G17. Partnership & 40 &  5\\ 
Total & 261 & 167 \\
\bottomrule
    \end{tabular}
    }
    \caption{Number of upsampled papers by SDGs. We first identify the candidate papers (\# Matched Papers), and then manually annotate them (\# Papers after Annotation). In total, there are 261 candidate papers, from which 167 are identified as relevant for NLP4SG.}
    \label{tab:upsample_details}
\end{table}

\subsubsection{Annotating Task 3: Salient Scientific Term Analysis}\label{appd:annot_3}

We present the title and abstract of social good papers to the annotator and ask them to write down a list of the primary tasks and methods addressed by each paper. In case there are hyponymy and hypernymy relationships among the scientific terms, we take the more specific term that is still common enough to show up in the Sankey diagram. For example, we would use the term LSTMs, but not neural networks (too broad), or stacked BiLSTMs with co-attention (too specific to be frequent enough to show up on the Sankey diagram). We utilized this information to enable the annotators to choose the relevant task and method spans from the title and abstract, along with an extra text box as an option in case an exact span was not available.

Below are the list of some common tasks and methods.

List of common tasks: automatic speech recognition, COVID-19-related analysis, text classification, data collection, event extraction, fact checking, fake news detection, gender bias mitigation, information extraction, information retrieval, linguistic analysis, machine translation, NLP applications, named entity recognition, natural language generation, question answering, relation extraction, rumor detection, sentiment analysis, stance detection, text summarization, toxicity detection, etc.

List of common methods: BERT, classifiers, computational models, conditional random fields, convolutional neural networks, ensemble methods, LSTM, multi-task learning, RoBERTa, supervised learning, support vector machines, topic models, transfer learning, word embeddings, etc.
In the absence of more specific methods, it is also acceptable to annotate more coarse-grained methods if they are the only ones mentioned in the paper: language models, language technology, machine learning methods, machine translation system, NLP models, neural networks, recurrent neural networks, transformers, etc.

\begin{table}[ht]
    \centering \small
    \begin{tabular}{p{7.3cm}lllll}
    \toprule
    \textbf{Title:} Applying deep learning on electronic health records in Swedish to \textbf{predict healthcare-associated infections} \\
    \textbf{Abstract:} Detecting healthcare-associated infections poses a major challenge in healthcare. Using natural language processing and machine learning applied on electronic patient records is one approach that has been shown to work. However the results indicate that there was room for improvement and therefore we have applied deep learning methods. Specifically we implemented a network of stacked sparse \textbf{autoencoders} and a network of stacked \textbf{restricted Boltzmann machines}. Our best results were obtained using the {stacked restricted Boltzmann machines} with a precision of 0.79 and a recall of 0.88. \\
    \\
    \textbf{Task:} predict healthcare-associated infections %
    \\
    \textbf{Method:} restricted Boltzmann machines, autoencoders
    \\
    \bottomrule
    \end{tabular}
    \caption{Example annotation of Task 3.}
    \label{tab:entity_extraction_example}
\end{table}

\subsection{Annotator Background}

The data associated with each task was annotated by two NLP researchers who are proficient in English. There are both female and male researchers, as well as white and Asian demographic backgrounds.

\subsection{Inter-Annotator Agreement}

For Task 2, we computed the Cohen's Kappa score of the binary classification of each SDG as shown in \cref{tab:annotator_agreement}, and calculate the weighted average as 88.67\%. Moreover, the partial match score is 96.46\% and the exact match is 85.35\%.

\begin{table}[ht]
    \centering\small
    \setlength\tabcolsep{2.7pt}
    {%
    \begin{tabular}{lcc}
\toprule
\textbf{UN SDG} & Cohen's kappa  \\ \hline
G1. Poverty & --  \\
G2. Hunger & --\\
G3. Health & 92.48\% \\
G4. Education & 95.19\% \\
G5. Gender & 86.32\% \\
G6. Water & --  \\
G7. Energy & -- \\
G8. Economy & 73.49\% \\
G9. Innovation & 84.92\% \\
G10. Inequalities &  68.96\%   \\
G11. Sustainable Cities &  33.15\%  \\
G12. Consumption & 100.00\%\\
G13. Climate & 100.00\% \\
G14. Life Below Water &-- \\
G15. Life on Land & -- \\
G16. Peace & 89.63\%   \\
G17. Partnership & 79.31\% \\ 
\bottomrule
    \end{tabular}
    }
    \caption{Cohen's kappa coefficients for each SDG.}
    \label{tab:annotator_agreement}
\end{table}

\section{More Dataset Information}

\subsection{License}
License or terms for use and/or distribution: The dataset is open-sourced with the Creative Commons Attribution Share Alike 4.0 license. We provide the ACL Anthology paper IDs, title, abstract, publication year, and acknowledgements section from ACL Anthology, with our annotations. Materials prior to 2016 in the ACL Anthology are licensed under the Creative Commons Attribution-NonCommercial-ShareAlike 3.0 International License, and materials published in or after 2016 are licensed on a Creative Commons Attribution 4.0 International License.

\subsection{NLP4SG through the years}
While papers addressing social problems in NLP have been predominantly published in recent years, it's important to include older papers in our analysis to get a complete picture of the growth of NLP4SG. 
Some examples of NLPSG papers for every decade are: 
\begin{itemize}
\item 1960s: Automated Processing of Medical {E}nglish.  \url{https://aclanthology.org/C69-1101}
\item 1970s: Powerful ideas in computational linquistics - Implications for problem solving, and education.    \url{https://aclanthology.org/P79-1028.pdf}
\item 1980s: Temporal Inferences in Medical Texts. \url{https://aclanthology.org/P85-1002}
\item 1990s: ANTHEM: advanced natural language interface for multilingual text generation in healthcare (LRE 62-007).   \url{https://aclanthology.org/1996.amta-1.27}
\item 2000s: Collaborative Annotation of Sign Language Data with Peer-to-Peer Technology. \url{https://aclanthology.org/L04-1280/}
\item 2010s: Detecting Offensive Tweets in Hindi-English Code-Switched Language. \url{https://aclanthology.org/W18-3504/}
\item 2020s: Gender Bias in Machine Translation. \url{https://aclanthology.org/2021.tacl-1.51/}
\end{itemize}

\subsection{Task 2 Co-Occurrence Matrix}

\begin{figure}[ht]
    \centering
    \includegraphics[width=\columnwidth]{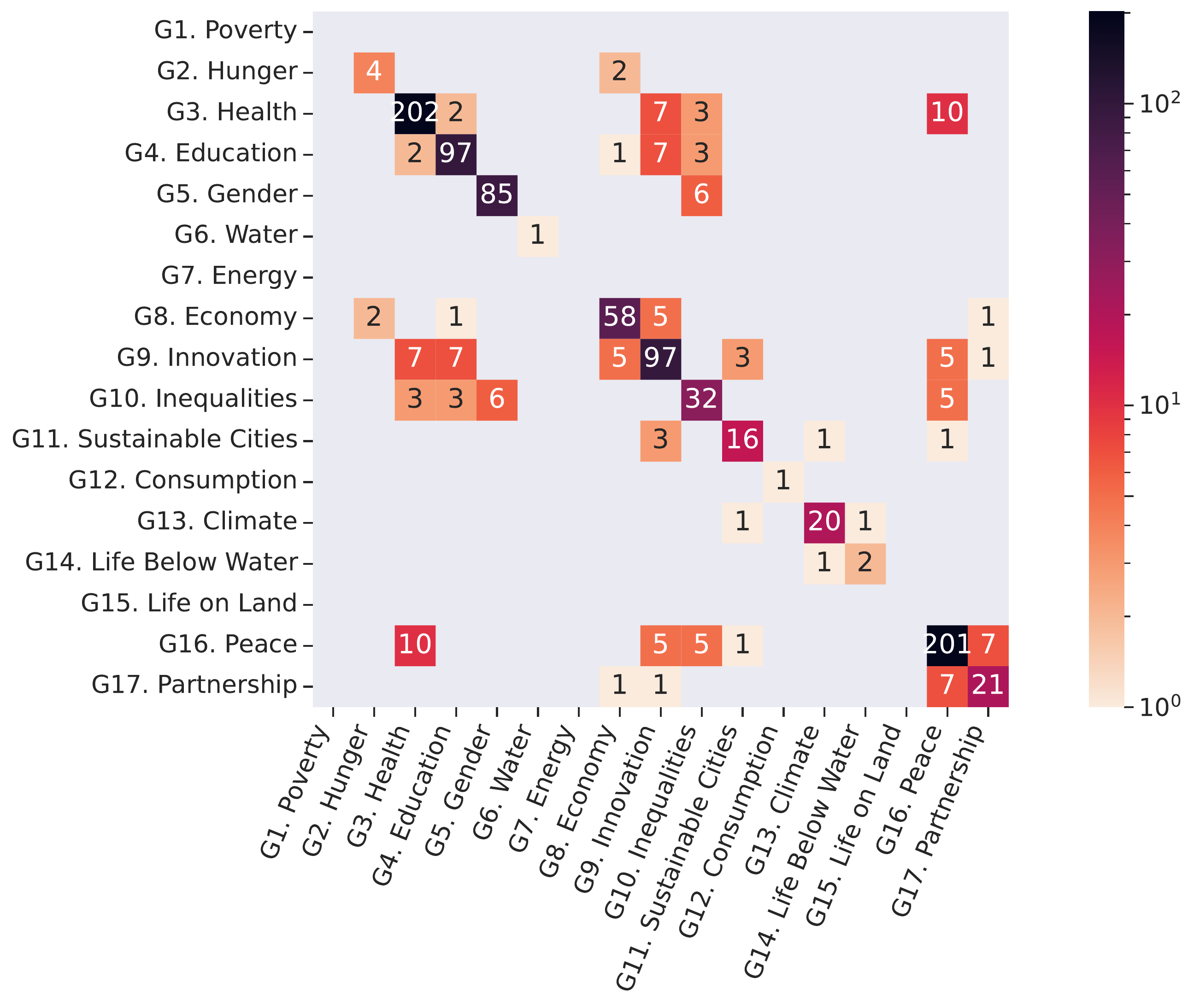}
    \caption{Heatmap of co-occurrences of UN SDGs on NLP4SG papers.}
    \label{fig:unsdg_heatmap_numbers}
\end{figure}

In \cref{fig:unsdg_heatmap_numbers}, we show the co-occurrence matrix with the absolute number of co-occurrences.

\section{Experimental Details}

\subsection{Models for Task 1}
\subsubsection{Distant Supervision}\label{appd:augment}
For the keyword matching, we curate a list of social good-related keywords 
\ifarxiv
provided in our GitHub repository
\else
in the codes uploaded to the submission system. 
\fi
which extends the AI for social good keywords by \citet{shi2020artificial}. We compare the performance of keyword matching based on the title vs. title and abstract in \cref{tab:distant_supervision}, and find that keyword matching in titles gives higher performance, so we adopt the title keyword matching as the first DS method.

For the second DS method, cosine similarity matching, we calculate
the cosine similarity scores between the text embeddings of the title + abstract of the paper and the text embeddings of the description of each one of the UN SDGs and take the most similar UN SDG. We add a paper to the positive set if its text similarity is among the top 1\% most similar papers in the unlabeled data from the ACL Anthology. We add a paper to the negative set if its similarity is in the bottom 50\% percentile. For the text embedding similarity, we use a general-purpose sentence similarity model, all-distilroberta-v1\footnote{\url{https://huggingface.co/sentence-transformers/all-distilroberta-v1}} which is a pre-trained distilroberta-base \citep{sanh2019distilbert} model fine-tuned on a 1B sentence pairs dataset.

Merging both DS sets, the entire DS set consists of positive samples from papers whose titles match the keywords, and papers which are very similar to the UN SDG descriptions; and negative samples from papers that are very different from the UN SDG descriptions.

We report the quality of data extracted by these DS methods evaluated on the training set in \cref{tab:distant_supervision},  which correlates with the ablation study results in \cref{tab:res_task1_ablation}. We can see that the performance of the entire DS set is higher than that of the keyword-only set, which is higher than the similarity-only set.

\begin{table}[ht]
    \centering \small
    \setlength\tabcolsep{2.7pt}
    \resizebox{\columnwidth}{!}{%
    \begin{tabular}{lcccccc}
    \toprule
& F1 & P & R  
\\ \midrule
    Keyword Match (Title) & 66.67 & \textbf{79.90} & 57.19 
    \\
    Keyword Match (Title+Abstract) & 55.17 & 42.50 & \textbf{78.60} 
    \\
    Cosine Similarity   & 8.33 & 47.06 & 4.57 
    \\
    Keyword Match (Title) + Cosine Similarity &  \textbf{72.02} & 76.85 & 67.76 
    \\
    \bottomrule
    \end{tabular}
    }
    \caption{Quality evaluation of the distant supervision heuristics reported on the training set. 
    }
    \label{tab:distant_supervision}
\end{table}

\begin{table}[ht]
    \centering \small
    \setlength\tabcolsep{4pt}
    \begin{tabular}{lcccc}
    \toprule
    & F1 & Acc & P & R \\ \midrule
   SciBERT with No DS & 73.42 & 94.10 & {84.90} & 64.68 \\
   SciBERT + DS by Similarity & 62.96 & 89.35 & 55.86 & 72.11 \\
   SciBERT + DS by Keyword  & 74.68  &  94.10 & 80.93  & 69.32\\
    \bottomrule
    \end{tabular}
    \caption{For the best-performing model, SciBERT+DS, we conduct ablation studies by removing the different subsets of our DS data. The resulting performance correlates with the DS data quality reported in \cref{tab:distant_supervision}, where keyword + similarity gets the best quality, and similarity only gets a very low quality.}
    \label{tab:res_task1_ablation}
\end{table}

\subsubsection{Models Implementation Details} \label{implem_details}\label{appd:model1_implementation}

\paragraph{BERT, RoBERTa, and SciBERT} We finetune SciBERT \citep{beltagy2019scibert} on the NLP4SG text classification task. We freeze the embedding layer and the first 11 encoder layers, and we finetune the parameters of the 12th encoder layer, the pooler, and the classifier layer. 

We use one GPU model NVIDIA Quadro RTX 6000 with 24 GiB.
We train the classifier for 15 epochs using a batch size of 32. The training time for our best-performing model is approximately 1.5 hours. We manually grid search with different learning rates (5e-03, 5e-04, 5e-05, 5e-06) and different scheduler types (``constant'',``linear'',``cosine with restarts'') to make a total of 12 runs and chose the best hyperparameters based on f1 score on the dev set. The chosen learning rate is 5e-05 and a linear scheduler. All the runs together make a total of around 18 GPU hours. We tried with a similar setting using BERT-large and RoBERTa-large, obtaining lower performances.

\paragraph{InstructGPT} For the second approach, we design a prompt and ask InstructGPT in a zero-shot setting if a paper is NLP4SG or not. The prompt contains 3 elements:
\begin{itemize}
\item Introduction to the task: We use the sentence ``There is an NLP paper with the title and abstract:''
\item Content of the paper: The title and abstract concatenated
\item Final question: We use the sentence ``Is this paper contributing to the UN Sustainable Development Goals? Answer yes or no.''
\end{itemize}

We use the OpenAI API\footnote{\url{https://beta.openai.com/overview}} to access GPT.
We use ``davinci-text-002.'' We keep the default values of the API, and set the temperature to zero and maximum number of tokens of the response as 50.

\paragraph{Few-shot InstructGPT}
We also perform experiments using few-shot settings, i.e., 2,4,8,16, and 32 shots, and report the results for the setting with the best performance in the dev set. For those experiments, we only use the title of the paper and the same proportion of positive and negative examples.  

\begin{quote}
Are these papers contributing to the UN Sustainable Development Goals or helping society?\\
Paper: \texttt{[Title + Abstract]}\\
Answer: \texttt{[Yes/No]}\\
Paper: \texttt{[Title + Abstract]}\\
Answer: \texttt{[Yes/No]}\\
Paper: \texttt{[Title + Abstract]}\\
Answer: \\
\texttt{[InstructGPT response]}\\
\end{quote}

\subsection{Models for Task 2}\label{appd:model2_implementation}

\paragraph{Model 1: InstructGPT }
We design a prompt to get which UN goal a paper contributes to. Since InstructGPT already has background knowledge of which are the UN SDGs, and what is the purpose of each one of them we do not need to include that explanation in the prompt. The prompt has the following components:
\begin{itemize}
\item Introduction to the task: We use the sentence ``There is an NLP paper with the title and abstract:''
\item Content of the paper: The title and abstract concatenated.
\item Final question: We use the sentence ``Which of the UN goals does this paper directly contribute to? Provide the goal number and name.''
\end{itemize}
We parse the response, allowing multiple goals for each paper.

\paragraph{Model 2: Pretrained MNLI classifier}
We use a pretrained MNLI classifier (bart-large-mnli) as a zero-shot text classifier, an approach proposed by \citet{yin-etal-2019-benchmarking}. We get the probability of entailment/contradiction of the content of the paper with the description of each one of the UN goals obtained from the UN website. We consider that a paper contributes to a UN SDG if the probability of entailment is 0.5 or higher. If there is no SDG with a probability higher than 0.5 we take the SDG with the highest probability. We also tested distilbert (typeform/distilbert-base-uncased-mnli) and DeBERTa (MoritzLaurer/DeBERTa-v3-base-mnli-fever-anli).

\subsection{Models for Task 3 }\label{appd:model3_implementation}

\paragraph{PURE and SciREX}
PURE \citep{zhong-chen-2021-frustratingly} is a model for end-to-end relation extraction. We use PURE trained on the SciERC \cite{luan-etal-2018-multi} dataset of scientific abstracts.

SciREX \citep{jain-etal-2020-scirex} is a document level IE model 
that not only identifies mentions, but also models their saliency, and their coreference links. 

For SciREX and PURE, we run them on a GPU model NVIDIA Quadro RTX 6000 with 24 GiB. It takes around one hour to extract the information from 9K social good papers.

\paragraph{BERT-based Models}
We used LLM fine-tunned on the SQUAD dataset \citep{rajpurkar2018squadrun}. We provide as context the title and abstract of the paper and asked ``Which NLP task does this paper address?'' and ``Which NLP method(s) does this paper use?''. In particular we used the following models ``bert-large-uncased-whole-word-masking-finetuned-squad'' , ``deepset/roberta-large-squad2'', and ``deepset/deberta-v3-large-squad2''.

\paragraph{InstructGPT}
We use InstructGPT in a zero-shot setting in order to extract the relevant information. We use a different prompt for each of the 2 extraction tasks (tasks and methods). The prompt consists of 3 elements:
\begin{itemize}
    \item Instruction: Here we tell InstructGPT what to do in natural language form.
    \item Paper content: We concatenate the title and abstract of the paper.
    \item Final question: We present the question again to reinforce the explanation.
\end{itemize}

\paragraph{Prompt to obtain the tasks}

\begin{quote}
Identify the NLP task(s) that this paper is dealing with. Select a text span that is an appropriate answer, or if no span serves as a good answer, just come up with a phrase. Examples of tasks are: fake news detection, name entity recognition, question answering, etc.\\
\texttt{[Title + Abstract]}\\
The primary NLP task(s) addressed in this paper are:\\
\texttt{[InstructGPT response]}\\
\end{quote}

\paragraph{Prompt to obtain the methods}

\begin{quote}
Identify the NLP method(s) used in this paper. Select a text span that is an appropriate answer, or if no span serves as a good answer, just come up with a phrase. Separate the methods with commas and don't include NLP tasks. Examples of methods are: BERT, SVM, CNN, etc.\\
\texttt{[Title + Abstract]}\\
The primary NLP method(s) used in this paper are:\\
\texttt{[InstructGPT response]}\\
\end{quote}

\section{Details of the Interpretability Analysis}

\subsection{Task 1: Example LIME Outputs}\label{ap_res_t1}
\begin{figure}[ht]
    \centering
    \ifshowfig
    \includegraphics[width=\columnwidth]{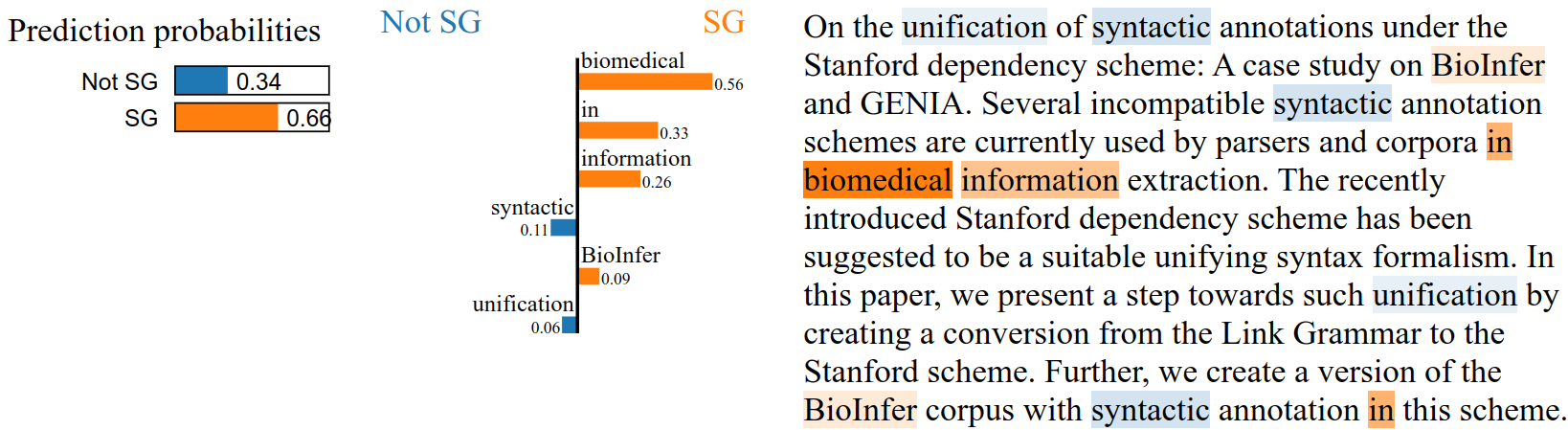}
    \fi
    \caption{LIME output for a TP example.}
    \label{fig:lime_tp}
\end{figure}

\begin{figure}[ht]
    \centering
    \ifshowfig
    \includegraphics[width=\columnwidth]{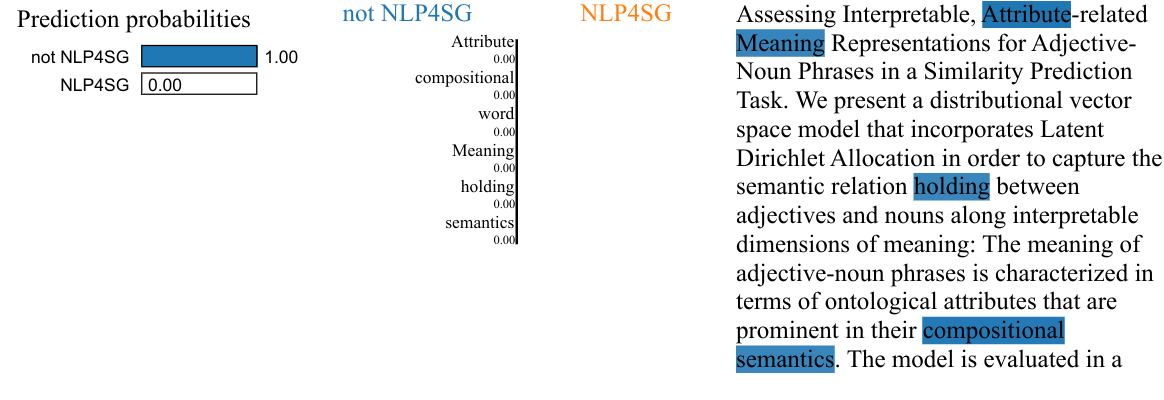}
    \fi
    \caption{LIME output for a TN example.}
    \label{fig:lime_tn}
\end{figure}

\begin{figure}[ht]
    \centering
    \ifshowfig
    \includegraphics[width=\columnwidth]{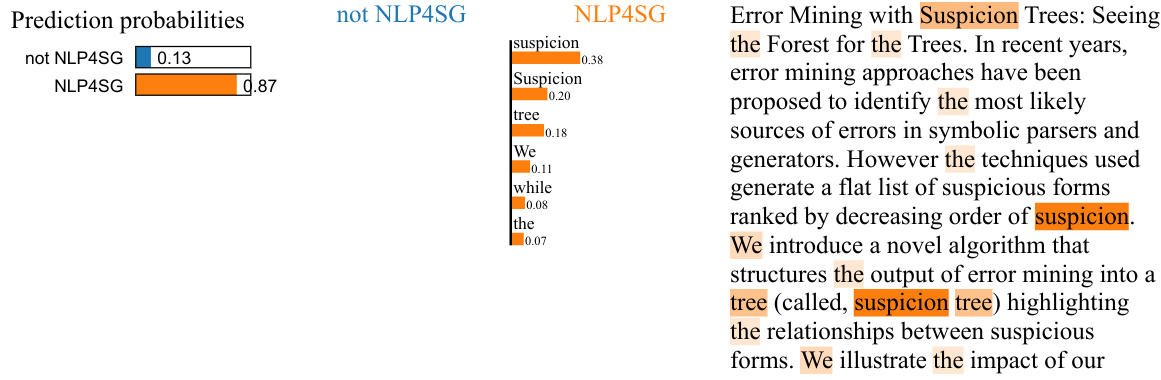}
    \fi
    \caption{LIME output for an FP example.}
    \label{fig:lime_fp}
\end{figure}

\begin{figure}[ht]
    \centering
    \ifshowfig
    \includegraphics[width=\columnwidth]{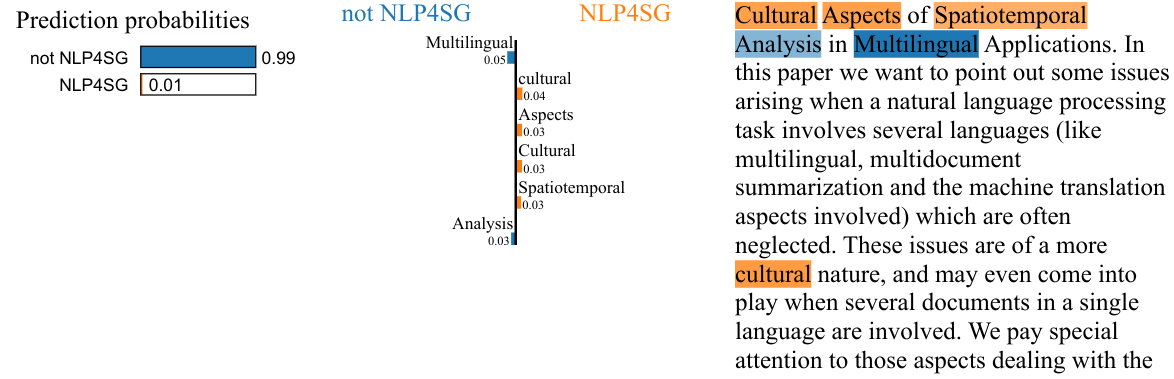}
    \fi
    \caption{LIME output for an FN example.}
    \label{fig:lime_fn}
\end{figure}

Using the best model, we show examples of LIME output for true positive (TP), true negative (TN), false positive (FP), and false negative (FN) data samples in \cref{fig:lime_tp,fig:lime_tn,fig:lime_fp,fig:lime_fn}, respectively.

\showfigtrue

\subsection{Task 2: Analyzing the Reasoning Chain of InstructGPT}\label{ap_res_t2} \label{appd:annot_explanation}
For the correctly classified samples by the best InstructGPT method, we perform further analysis to check the \textit{explanation quality} of the model. 

We show two motivating examples in \cref{tab:llm_understanding_abrupt}, where the first example shows sufficient intermediate reasoning steps starting from what the paper contributes, to intermediate effects of the paper, and finally to the end social impact, while the second example lacks intermediate reasoning steps, and mentions only what the paper does, followed the UN SDG right after, without intermediate explanations.

Specifically, we denote a good ``impact chain'' as a complete and cohesive reasoning chain from the direct description of the paper's research output to intermediate impact, and end impact. Then we randomly select 200 test samples, and use the following two criteria to check the explanation quality of InstrutGPT: (1) whether the mentioned ``impact chain(s)'' sound cohesive and reasonable, e.g., not missing necessary intermediate reasoning steps, which is analogous to the precision score, and (2) whether there are other obvious impact chains that are missing, which is analogous to the recall score. For each criterion, we assign 1 for good performance, 0 for bad performance, and 0.5 for partially good performance. 

We find that 77.6\% of the explanations contain cohesive and convincing reasoning chains, and 
18\%
of the explanations miss at least one obvious impact chain of how the paper contributes to UN SDGs.

\paragraph{SDG classification explanation}
\begin{quote}
There is an NLP paper with the title and abstract:\\
\texttt{[Title + Abstract]}\\
Is this paper contributing to the UN Sustainable Development Goals? Answer yes or no. If the answer is ``yes'', mention which goal the paper is contributing to and in which way it is contributing to that goal.\\
\texttt{[InstructGPT response]}\\
\end{quote}

\begin{table}[ht]
    \centering \small
    \begin{tabular}{p{7.3cm}lllll}
    \toprule
    \textit{\textbf{An Example with Sufficient Reasoning Steps for the Impact Chain}}
    \\
    \textbf{Title:} Extracting Symptoms and their Status from Clinical Conversations \\
    \textbf{Abstract:} This paper describes novel models tailored for a new application, that of extracting the symptoms mentioned in clinical conversations along with their status. 
    [...]
    We propose two novel deep learning approaches to infer the symptom names and their status: 
    [...]
    This task stems from a realistic application of assisting medical providers in capturing symptoms mentioned by patients from their clinical conversations. 
    [...]
    \\
    \textbf{InstructGPT Response:} Yes, this paper is contributing to the UN Sustainable Development Goal of \textit{Good Health and Well-Being}. The paper describes models that are designed to \textit{extract symptoms from clinical conversations}, which can \textit{assist medical providers} in capturing symptoms mentioned by patients. This can ultimately lead to \textit{better diagnosis and treatment of diseases}, and \textit{improve overall health and well-being}.
    \\
    \midrule
    \textit{\textbf{An Example Lacking Intermediate Reasoning Steps for the Impact Chain}}
    \\
    \textbf{Title:} Extracting Symptoms and their Status from Clinical Conversations \\
    \textbf{Abstract:} Lexically-Triggered Hidden Markov Models for Clinical Document Coding. The automatic coding of clinical documents is an important task for today's healthcare providers. Though it can be viewed as multi-label document classification, the coding problem has the interesting property that most code assignments can be supported by a single phrase found in the input document. We propose a Lexically-Triggered Hidden Markov Model (LT-HMM) that leverages these phrases to improve coding accuracy. 
    [...]
    \\
    \textbf{InstructGPT Response:} Yes, this paper is contributing to the UN Sustainable Development Goal of \textit{good health and well-being}. The paper is doing this by \textit{proposing a Lexically-Triggered Hidden Markov Model (LT-HMM)} that leverages phrases to \textit{improve coding accuracy} for clinical documents. This will \textit{ultimately help healthcare providers} provide better care to patients. 
    \\
    \bottomrule
    \end{tabular}
    \caption{Examples of sufficient and insufficient reasoning about the impact chain of NLP4SG papers.}
    \label{tab:llm_understanding_abrupt}
\end{table}

\section{More Insights}

\subsection{Survey}\label{appd:survey}
\begin{figure}[ht]
    \centering
    \ifshowfig
    \includegraphics[width=\columnwidth]{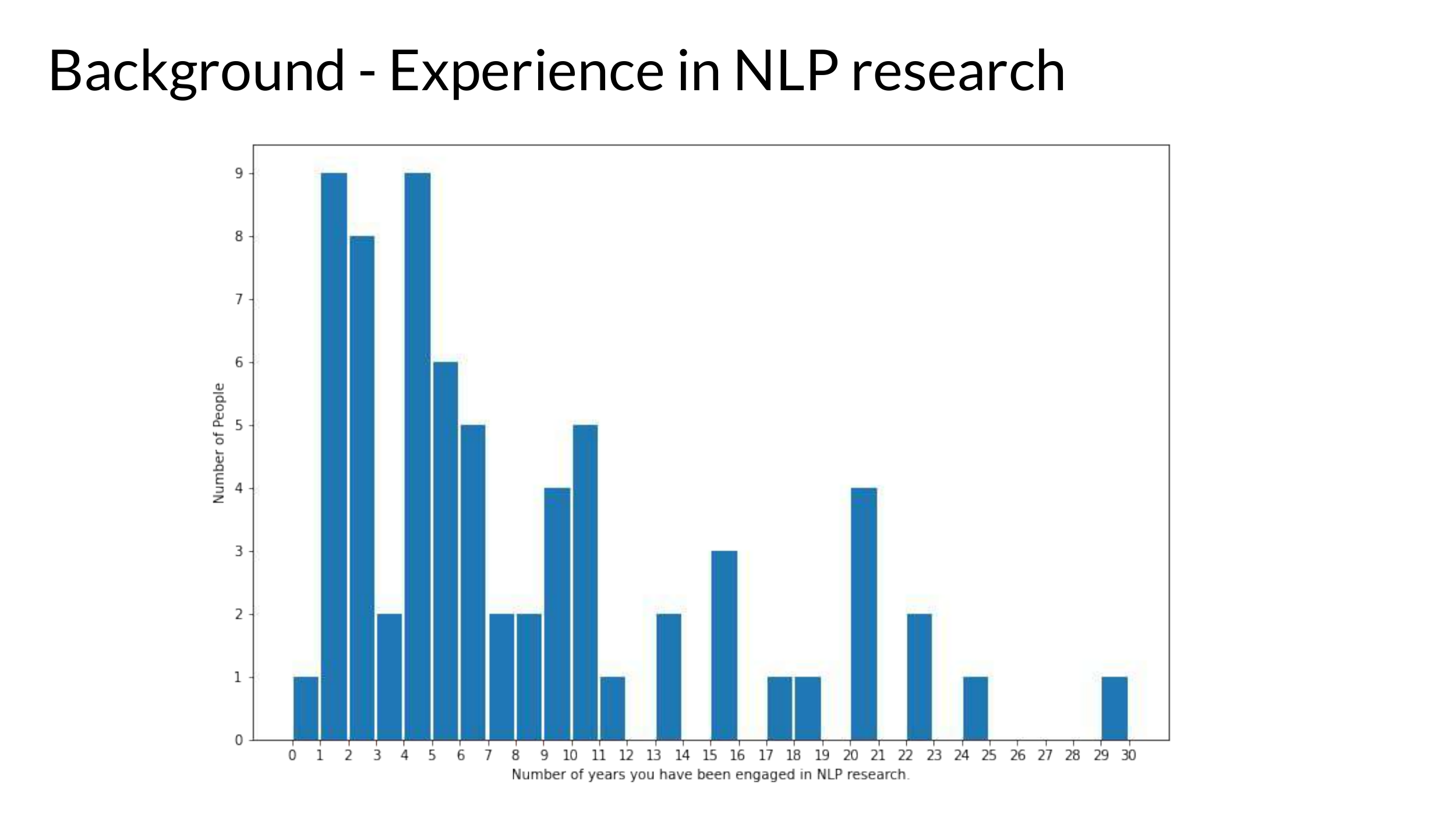}
    \includegraphics[width=\columnwidth]{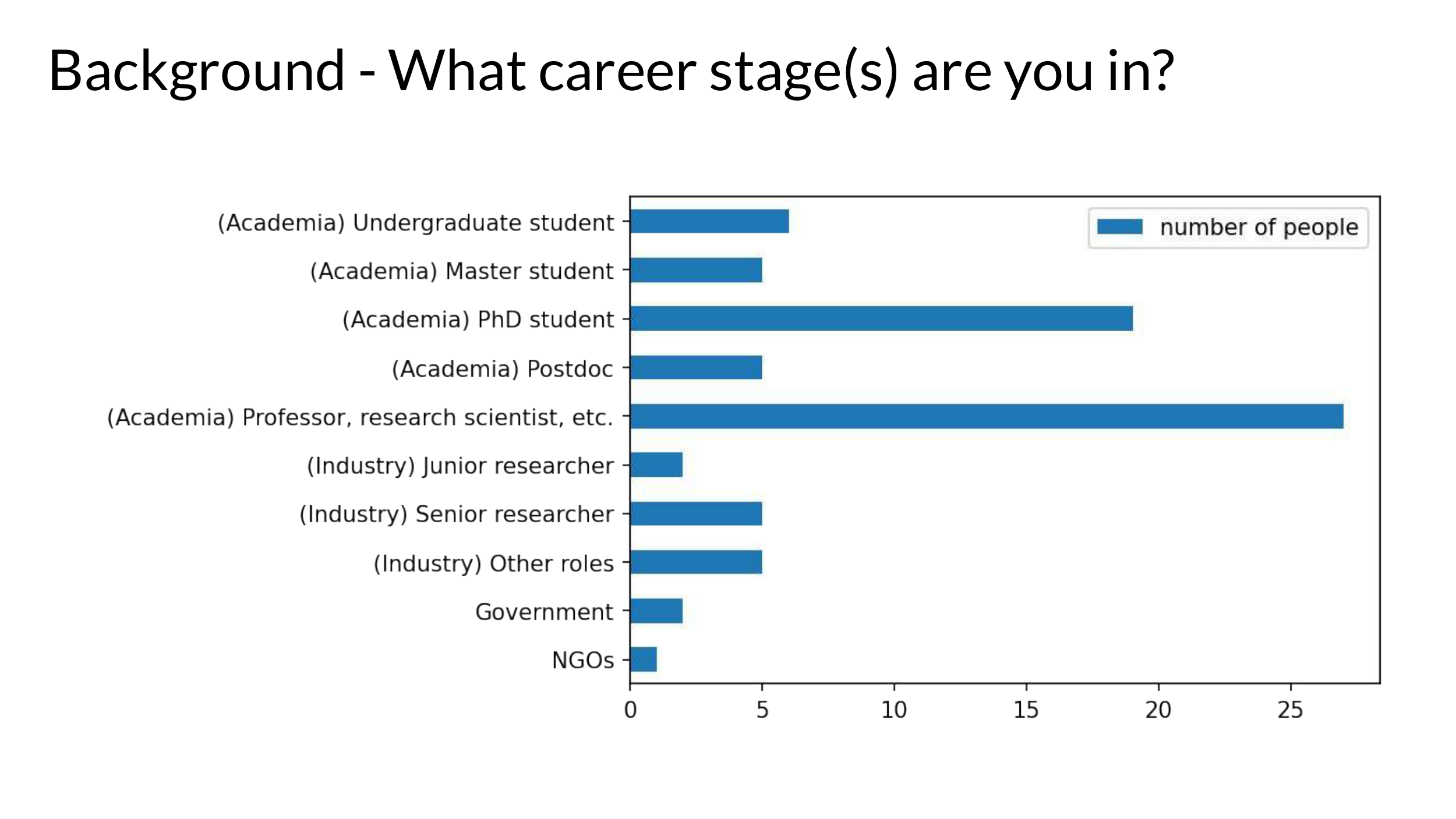}
    \includegraphics[width=\columnwidth]{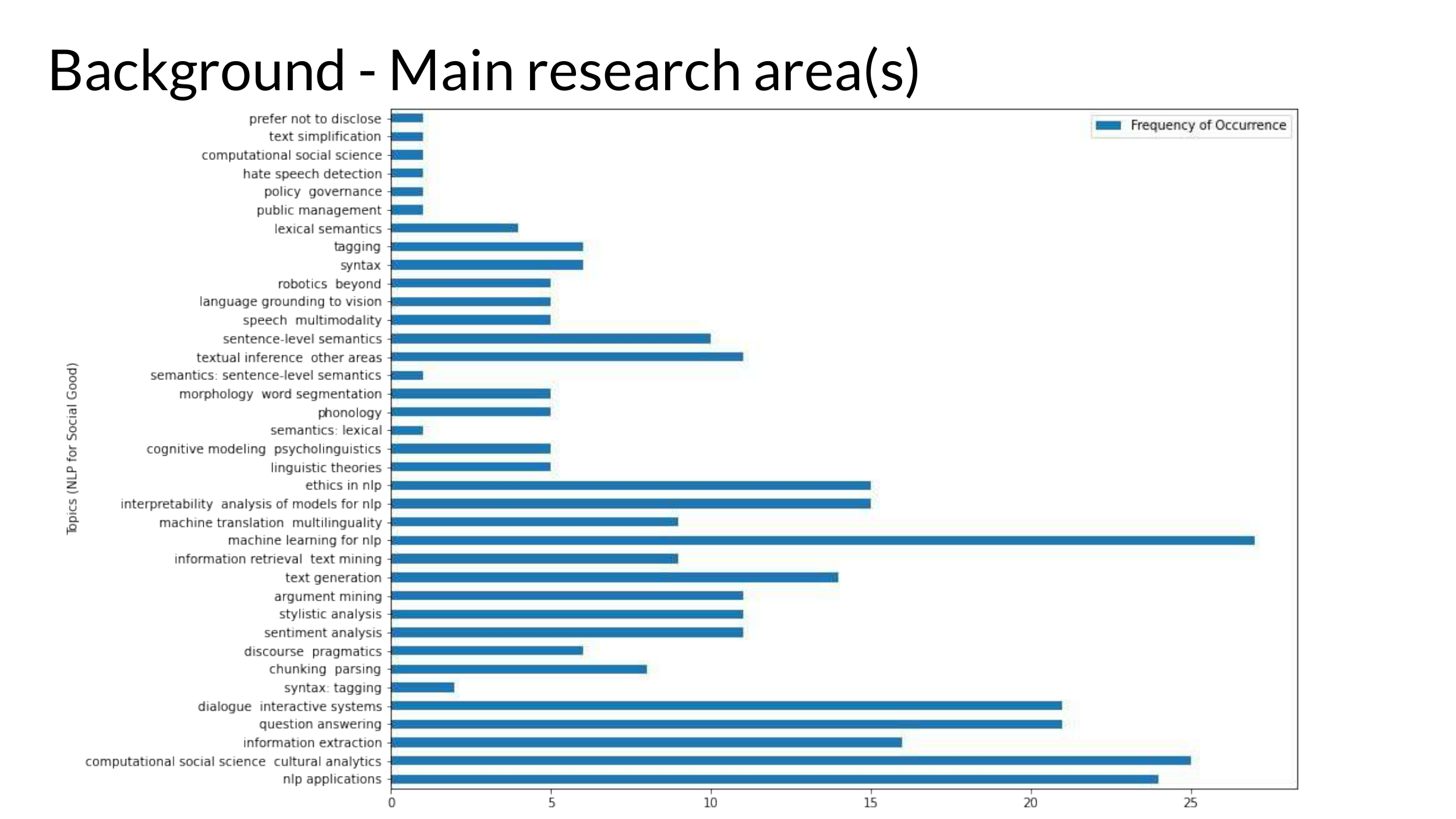}
    \fi
    \caption{Backgrounds of the survey subjects, including the number of years in NLP research, career stage, and research areas.
    }
    \label{fig:survey}
\end{figure}

We conduct a survey from July 31, 2021 to August 31, 2021
about ``How current NLP researchers think about NLP4SG.'' We receive 
80 
responses from NLP researchers, among which more than 30\% are NLP professors. 
\ifarxiv
We reach the subjects at the ACL conference and also through emails.
\fi
The demographics of the survey subjects are shown in \cref{fig:survey}.

The question in the survey that contributes to the SDG visualization is ``What social good applications do you think the NLP community should consider?'', for which we manually parse the open-ended answers to corresponding SDGs. Example answers are 
NLP for Education (22.9\%),
physical and mental health (18.6\%), 
and combating misinformation (17.1\%).

\subsection{Ideal SDG alignment we want to push for }\label{appd:discuss_alignment}

\begin{figure}[ht]
    \centering
    \includegraphics[width=\columnwidth]{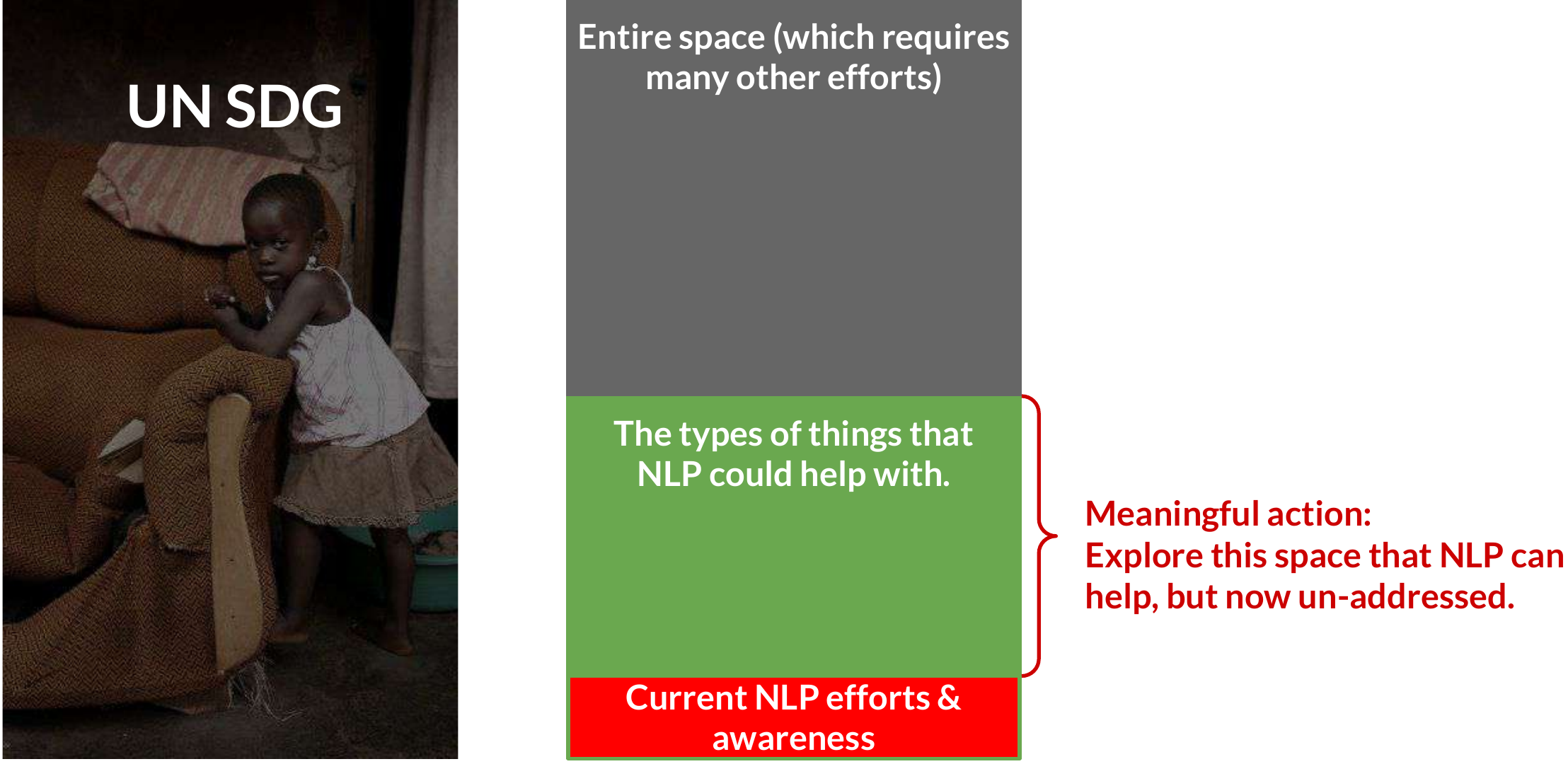}
    \caption{Illustration of the action space NLP researchers can aim for. }
    \label{fig:insight_action_space}
\end{figure}

We show an ideal framework in \cref{fig:insight_action_space}, where for each goal, NLP can help with a subset of the things. And the meaningful action space is the gap between the current NLP research landscape vs. what NLP can potentially do.

\section{More Website Details} 

\subsection{Implementation Details} \label{appd:website_implementation}
Although we run the \ourmodel on the entire ACL Anthology. For visualization, we mostly visualize 1980 to 2021, because there were not many papers before 1980, and also the progress tracking of NLP4SG in the recent decades might be more relevant for the audience. Also, since our data was crawled in the middle of 2022, the trend of 2022 is not comparable to the previous year 2021, so we omit 2022 for the moment, but it could be added in the future.

For the visualization of the Sankey diagram, we conduct some preprocessing over the salient scientific terms produced by the model.
Since there are variances in the expression of the same task or method, such as ``long short-term memory networks,'' and ``LSTMs'', we group similar 
terms together with the following procedure:
\begin{itemize}
    \item We get a vector representation for each word with Word2Vec \citep{word2vec}. We train Word2Vec with the abstracts of all the ACL Anthology papers, we set the parameter window to 10 and vector size 100.
    \item We use the agglomerative clustering function in scikit-learn \citep{scikit-learn} on the word vectors, with the parameter distance threshold set to 0.2.
    \item We measure the  ratio score similarity between each word of a group to its centroid and just keep words in the cluster that have a similarity score greater than 75. We used the library "TheFuzz"\footnote{\url{https://github.com/seatgeek/thefuzz}} which uses Levenshtein distance to compute its similarity measure.
    \item We take that automatic clustering as a starting point and then manually check each one of the clusters, add new elements and create new clusters if necessary.
\end{itemize}

\subsection{Demo on the Gold Data}\label{appd:gold_demo}

As mentioned in the Limitations section, one constraint is that the visualizations are bottle-necked by the model performance. As a supplementary, we also provide the visualization based on our annotated \ourdata dataset, which is smaller but accurate.

\begin{figure}[ht]
    \centering
    \includegraphics[width=\columnwidth]{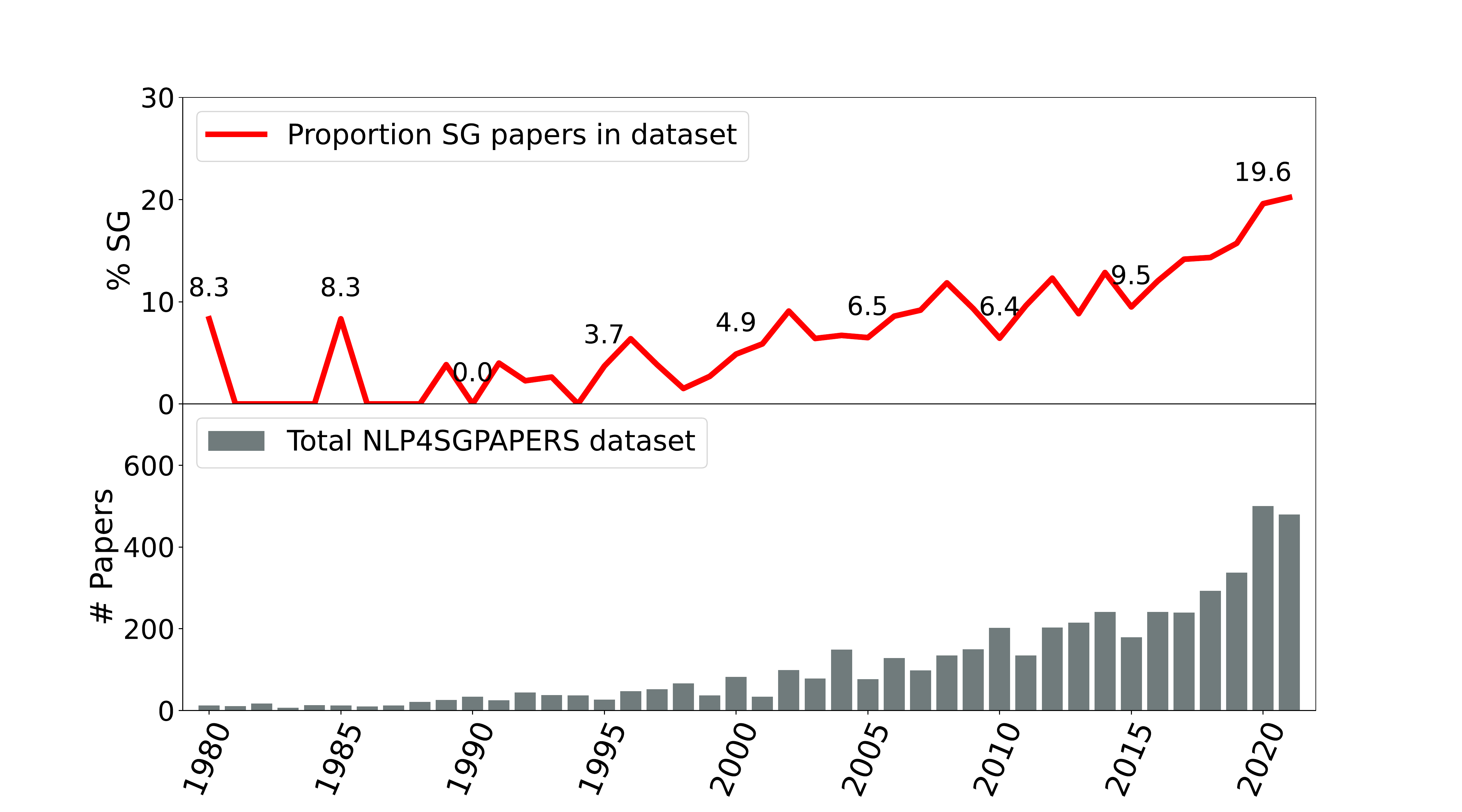}
    \caption{Percentage of social good-related papers (top) and the total number of papers (bottom) in \ourdata each year from 1980. }
    \label{fig:dist_papers_dataset}
\end{figure}

We produce all the visualization figures mentioned in \cref{sec:insights}, but based on the ground-truth annotation in our \ourdata dataset. \cref{fig:dist_papers_dataset} visualizes the proportion of NLP4SG papers over the years, where  the proportion of NLP4SG papers steadily increased from 2000.

\begin{figure}[ht]
    \centering
    \ifshowfig
    \includegraphics[width=\columnwidth]{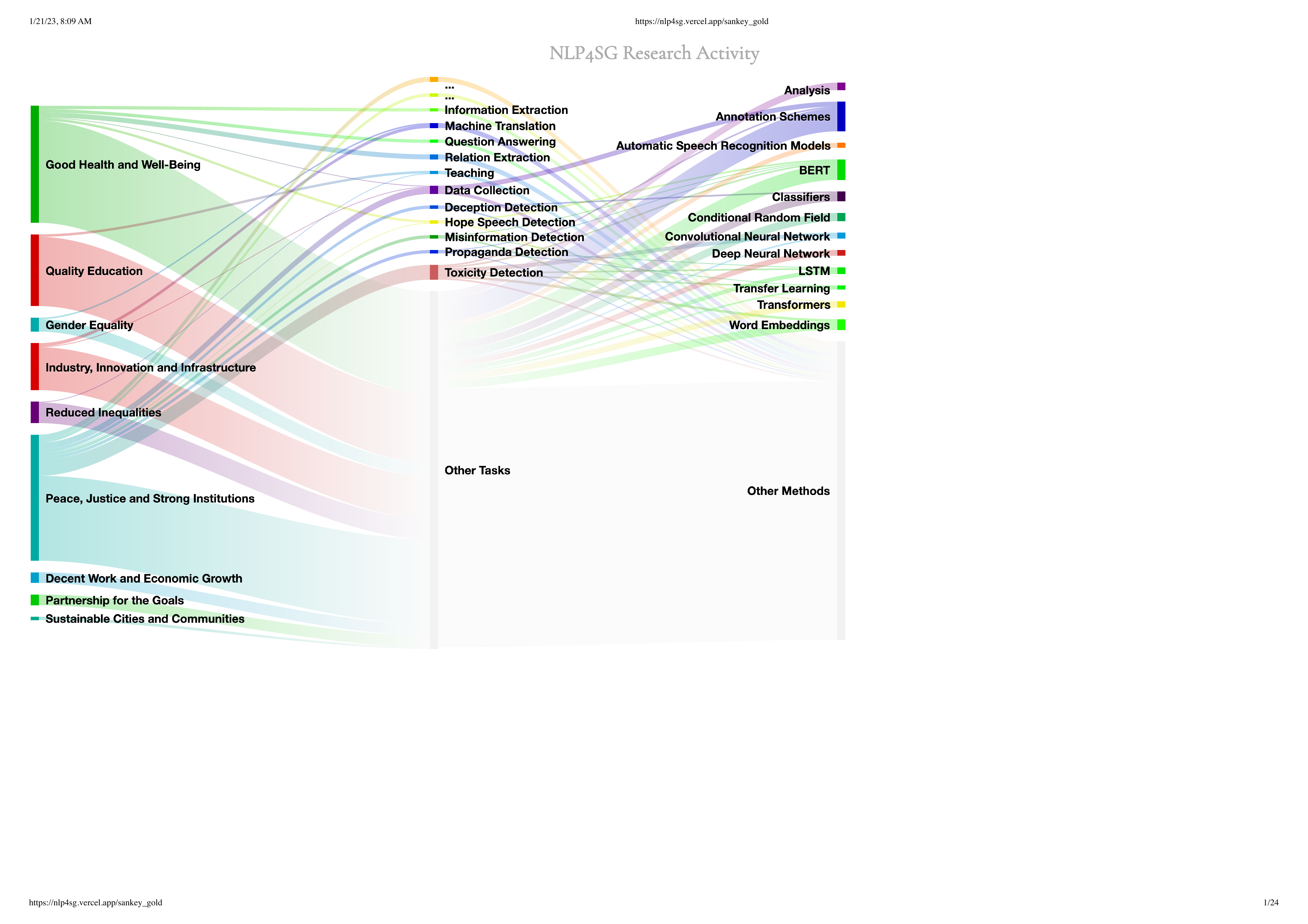}
    \fi
    \caption{Sankey diagram based on gold 
    \ifarxiv
    annotations at \url{https://nlp4sg.vercel.app/sankey_gold}.
    \else
    annotations.
    \fi
    }
    \label{fig:website_sankey_gold}
\end{figure}

\begin{figure}[ht]
    \centering
    \includegraphics[width=\columnwidth]{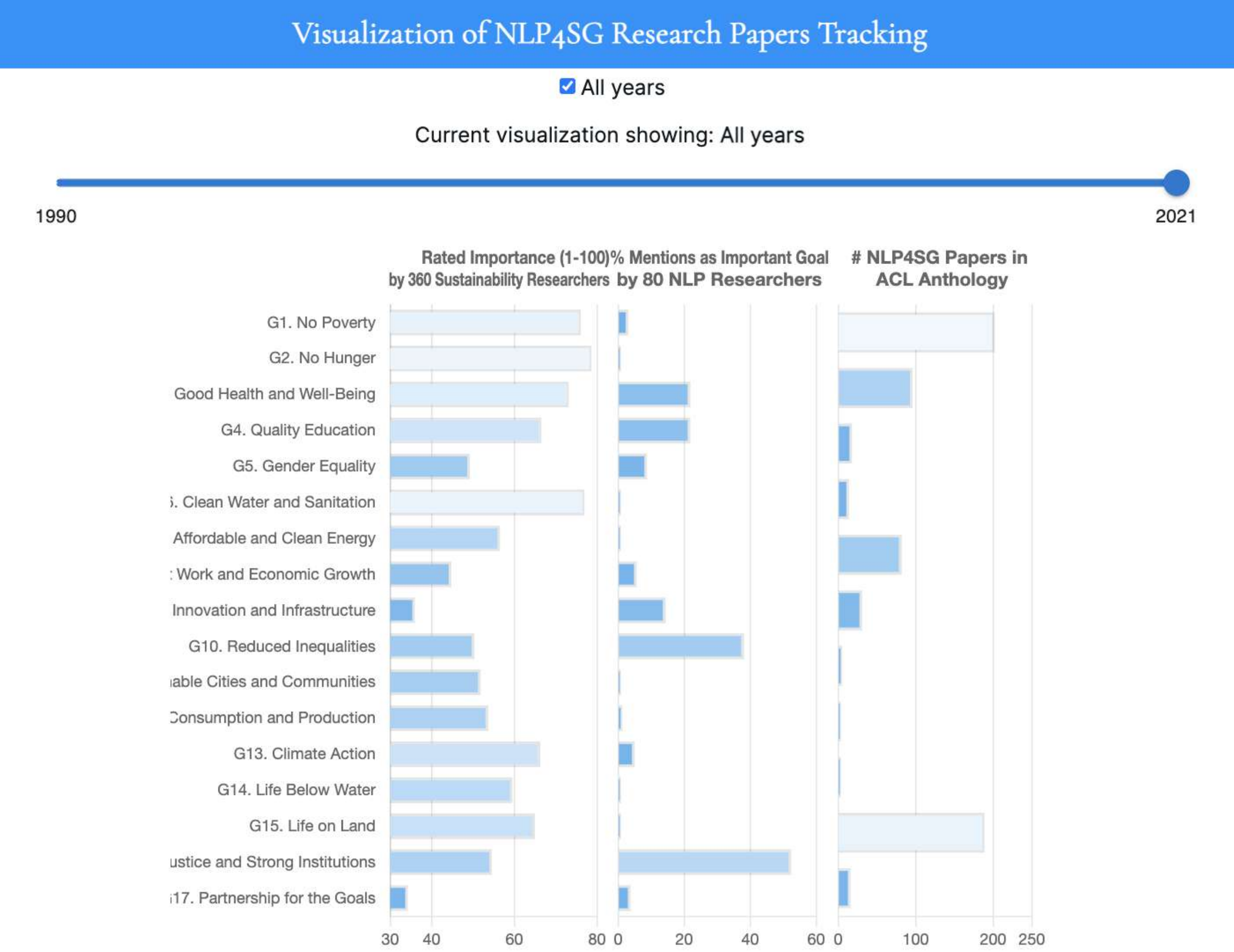}
    \caption{SDG visualization based on gold 
    \ifarxiv
    annotations at \url{https://nlp4sg.vercel.app/sdg_gold}.
    \else
    annotations.
    \fi
    }
    \label{fig:website_sdg_gold}
\end{figure}

We also show the Sankey diagram, using the gold data only, in \cref{fig:website_sankey_gold} and the SDG visualization in \cref{fig:website_sdg_gold}.

\end{document}